\documentclass{article}

\usepackage[numbers]{natbib}

\usepackage {arxiv}
\usepackage{cite}
\usepackage{floatrow}
\usepackage{amsmath,amssymb,amsfonts}
\usepackage{caption}
\usepackage{algorithm}
\usepackage{algorithmic}
\usepackage{textcomp}
\usepackage[table]{xcolor}
\usepackage{float}
\usepackage{booktabs}
\usepackage{enumitem}
\usepackage{setspace}
\usepackage[]{graphicx}
\usepackage{amsmath}
\usepackage{amssymb}
\usepackage{epsfig}
\usepackage{epstopdf}

\usepackage{subfigure}
\usepackage{caption}
\usepackage{comment}
\usepackage{multirow}

\title{Perturbation of Deep Autoencoder Weights for Model Compression and Classification of Tabular Data}

\author{Manar D. Samad and Sakib Abrar\\
Department of Computer Science\\
 Tennessee State University\\
 Nashville, TN, USA\\
\texttt{msamad@tnstate.edu} 
}

\begin{document}

\maketitle
\begin{abstract}
Fully connected deep neural networks (DNN) often include redundant weights leading to overfitting and high memory requirements. Additionally, the performance of DNN is often challenged by traditional machine learning models in tabular data classification. In this paper, we propose periodical perturbations (prune and regrow) of DNN weights, especially at the self-supervised pre-training stage of deep autoencoders. The proposed weight perturbation strategy outperforms dropout learning in four out of six tabular data sets in downstream classification tasks. The L1 or L2 regularization of weights at the same pretraining stage results in inferior classification performance compared to dropout or our weight perturbation routine. Unlike dropout learning, the proposed weight perturbation routine additionally achieves 15\% to 40\% sparsity across six tabular data sets for the compression of deep pretrained models. Our experiments reveal that a pretrained deep autoencoder with weight perturbation or dropout can outperform traditional machine learning in tabular data classification when fully connected DNN fails miserably. However, traditional machine learning models appear superior to any deep models when a tabular data set contains uncorrelated variables. Therefore, the success of deep models can be attributed to the inevitable presence of correlated variables in real-world data sets.   
\end {abstract}



\section{INTRODUCTION}

The recent advancement in artificial intelligence is primarily attributed to the deep learning of image, video, and audio data~\citep{ALAM2020}. The majority of deep learning studies propose different variants of convolutional neural networks (CNNs) to demonstrate image classification performance. The prevalence of CNN is because of the convolution operation within neural networks that can extract highly discriminating patterns in data with spatial regularity (e.g., images, audio). Furthermore, the hierarchical patterns in images are sequentially captured in multiple layers of CNN, where the final convolutional layer yields the most semantic features for image classification. These achievements of CNN overshadow the performance of deep learning on data without such spatial regularity or hierarchical patterns. The data mining literature broadly categorizes data into three types: 1) imaging data, 2) text data, and 3) tabular data ~\citep{Zhang_elsevier_2018, Mueller_ACM_2020}. Tabular data often do not contain hierarchical patterns or spatial or temporal regularity to effectively leverage the feature extraction mechanisms of popular convolutional or recurrent neural networks (RNN)~\citep{IMP_Ucar_neurips_2021, IMP_Somepalli_arXiv_2021}. In general, tabular data sets have received much less attention in the deep learning literature than imaging or text data. There are studies reporting the superiority of traditional ensemble learning methods, such as gradient boosting trees, over deep learning approaches in mining tabular data sets~\citep{popov_arXiv_2019,Hatwell_Springer_2020}. One example of tabular data is electronic medical records (EMR), a structured collection of patients' vitals and lab measurements. Our experience with large EMR data sets reveals that traditional ensembles of decision tree classifiers outperform FC-DNN in all-cause mortality prediction~\citep{Samad2018}. This suboptimal performance of FC-DNN is burdened with high computational cost, lack of model explainability, and large memory requirements, which deter many domain researchers (e.g., health scientists) from effectively utilizing the true strengths of deep learning. 

The suboptimal performance of deep learning on tabular data may be due to 1) poor feature representations and statistical assumptions on variables, 2) overfitting due to flexibility of deep learning with limited sample size, and 3) redundancy in deep architecture.  First, an FC-DNN can be pretrained as a self-supervised autoencoder to enhance the feature representation without involving classification labels. The pre-trained model is then fine-tuned to perform a target classification task. This transfer learning approach is known to be comparatively more effective than training FC-DNN from scratch. Second, the overfitting problem of neural networks is tackled using dropout learning and weight regularization~\citep{srivastava_jmlr_2014,poernomo2018biased}. Third, the redundancy in fully connected deep models is lessened via numerous weight pruning techniques, creating a sparse model. A unified framework that combines the objectives of feature learning, model regularization, and compression to augment the performance of deep learning of tabular data can be valuable.       

This paper proposes an algorithm to periodically perturb weights of deep autoencoders to simultaneously yield compressed pretrained models and improved feature representation for downstream classification tasks. The proposed perturbation routine involves periodic pruning and growing of weights at the time of self-supervised model pretraining. It is noteworthy that standard weight pruning methods are performed monotonically, conceding some model accuracy loss at the expense of introducing model compression. In contrast, we hypothesize that the perturbation of selective weights rather than randomly dropping neurons (dropout) or pruning weights can improve the downstream classification performance of deep models.

The remainder of the article is organized as follows. Section II provides a literature review on autoencoders, dropout learning, and weight pruning with a mathematical realization of autoencoder-based learning. Section III discusses the proposed algorithm, experimental setup, and model evaluation criteria. Section IV provides the findings following the proposed hypotheses and experimentation. Section V highlights the major findings with explanations and limitations of our work. The conclusions of this study are summarized in Section VI.

\section {Literature review}

A general trend in the deep learning literature is to build deeper and wider learning networks with thousands to millions of trainable parameters for achieving state-of-the-art performance. Therefore, some model accuracy can be gained at the cost of substantial computational and memory overheads at both the training and testing phases. Conversely, a deeper model is known to be more prone to overfitting. It is argued that most of these computing and memory requirements are redundant, which hints at the need for optimal network architectures~\citep{sehwag_arxiv_2019, liu_ICCV_2017}. In a medical image classification study using deep CNN, only eight out of 256 (3\%) image filters at the final convolutional layer have yielded non-zero filter responses~\citep{ICHI2021}. In other words, more than 96\% of the trained filters at the final feature extraction layer appear redundant in this particular scenario, consuming substantial CPU cycles and memory spaces. Therefore, reducing model complexity is imperative to 1) alleviate the notorious overfitting problem with deep learning and 2) introduce model compression for resource-restricted computing platforms. The reduction in model complexity can be achieved by dropping the redundant neurons (nodes) or the connection between neurons (weights).

\subsection{Dropping of neurons}
The dropout of neurons is commonly used to mitigate overfitting at the time of training a deep model~\citep{Lee_springer_2020}. In dropout, a percentage of the total neurons are deactivated to zero to discontinue their contributions to the next layer in training neural networks.  The random dropping of neurons at each training epoch improves the model's generalizability when tested on unseen test data. However, the random selection of neurons does not explain the importance of individual nodes to justify their inclusions or exclusions. In other words, the selection of neurons is not informed by any state or behavior of the network. Conversely, Hou et al. rate individual channels (filter responses) at each layer of a deep CNN using channel activation responses. This rating is used to identify and drop a subset of channels in the next layer~\citep{Hou_AAAI_2019}. Shen et al. have proposed a continuous dropout method that uses probability values instead of binary decisions to deactivate the neurons~\citep{Shen2018}. Salehinjad and Valaee implement dropout by learning a set of binary pruning state vectors using an energy loss function. Their dropout achieves more than 50\% model compression at the loss of less than 5\% of top-1 classification accuracy~\citep{Salehinejad_IEEE_2021}. Pham et al. have introduced 'AutoDropout' to automate the process of dropout at all channels and layers of deep CNN~\citep{Pham_AAAI_2021}. AutoDropout learns the best dropout pattern to train a target network. In another study, Kolbeinsson et al. perform tensor factorization of tensor regression layers of CNN on which they apply dropout for image classification tasks~\citep{Kolbeinsson_IEEE_2021}. These studies suggest the general trend that proposed new variants of dropout learning are tested on image classification tasks using deep CNN.

\subsection{Dropping of weights}

The weights of deep neural networks are pruned to disconnect redundant connections between two neurons after training (static), or during training (dynamic)~\citep{Liang2021Survey,  jiang2021learning}.
Weights are commonly pruned based on their magnitudes, which is known as magnitude-based pruning. However, determining an optimal threshold for the magnitude is an open problem. Li et al. have proposed an optimization algorithm to determine the threshold on weight magnitude for layer-wise weight pruning~\citep{Li_IJCAI_2018}. Han et al. prune the weights of an already trained network and then finetune the remaining weights~\citep{Han_neurips_2015}. In contrast to magnitude-based pruning, Park et al. have proposed a score-based pruning method,  considering the relationship between two consecutive layers in CNN and transforming pruning into an optimization problem~\citep{Park_ICLR_2020}. All these pruning methods are aimed at pruning only or sparsifying deep neural networks to enable their usage in resource restricted platforms. 

A more recent approach is to prune and regrow weights to gain back some of the lost model complexity due to pruning. 
Guo et al. term this as 'dynamic network surgery' that prunes parameters during training and splices some of those pruned weights back to retain model accuracy~\citep{Guo_neurips_2016}. Lin et al. calculate the weight gradient on pruned weights, which is then used to update the weights in back propagation~\citep{Lin_ICLR_2020}. The feedback from the weight gradient prevents the premature elimination of weights in image classification tasks. He et al. periodically prune image filters following each training epoch but allow for updating the filters subsequently~\citep{He_ACM_2018}. Bellec et al. propose a deep rewiring method to periodically prune and regrow the weights during training~\citep{Bellec_ICLR_2018}. In contrast to zero initialization of pruned weights, Mocanu et al. prune weights at the end of each epoch but initialize them randomly to regrow subsequently~\citep{Mocanu_nature_2018}.

We identify several limitations of existing studies. First, all the proposed weight pruning, pruning-and-growing methods are applied to CNN models and image filters using benchmark imaging datasets, such as MNIST, CIFAR, and ImageNet. The terms 'DNN' and 'CNN' are often used interchangeably in the literature. There is little or no research on how these methods perform on fully connected deep neural networks with tabular datasets. Second, all pruning methods are reported and expected to incur some accuracy loss due to model compression. Third, Hookers et al. have recently demonstrated that a pruned neural network model is more sensitive to a subset of samples and vulnerable to adversarial attacks~\citep{Hookers}.

\subsection {Deep autoencoders}

Deep autoencoders are multi-layer neural networks designed to encode input data into a latent feature representation by self-supervised learning. The self-supervised objective function aims to decode the original input data from the latent feature representation. Therefore, it can serve multiple purposes in machine learning. First, the model can be used to reconstruct the actual data from corrupt data inputs~\citep{eraslan_nature_2019}. Second, the latent feature representation of autoencoders may capture the data manifold in low dimensions for subsequent classification or clustering tasks. Third, the self-supervised mechanism of autoencoders alleviates the need for data labels for feature extraction because data labels are often expensive to acquire. Fourth, autoencoders can pretrain a DNN in a self-supervised way using a large data set without data labels. The pre-trained model can be fine-tuned for a given classification task with limited data labels. Therefore, an autoencoder enables a more effective transfer of knowledge and weight initialization than training a deep model from scratch. Similar to other machine learning models, autoencoders are mainly studied with image data, such as image reconstruction\citep{li_neucom_2021,ranjan_cvf_2018}, and image denoising\citep{Zheng_TIME_SERIES_2022, eraslan_nature_2019}. Autoencoders are proposed in studies on natural language generation, translation, and comprehension using text data sets~\citep{lewis_acl_2019}. \

\subsection{Contributions}
In reference to the literature review, the contributions of this paper are as follows. First, this study is one of the first to extensively evaluate the effect of weight pruning on deep autoencoders in learning tabular data sets. Second, a new weight perturbation routine is proposed that periodically performs weight pruning and regrowing at the deep model's pretraining stage. Third, the weight perturbation routine is modeled to simultaneously achieve model compression and improved feature learning instead of trading one for the other. Fourth, weight perturbation is compared against popular regularization of neurons and weights  (e.g., dropout, weight regularization) using high-dimensional tabular data sets from six research domains. Fifth, we compare the classification performances between deep (with and without perturbation) and traditional machine learning to explain when deep learning of tabular data sets fails.

\section {Background review}

Autoencoders use a self-supervised mechanism to first encode (encoder) input data (X$\in \Re^d$) into a new feature embedding (Z$\in\Re^m$) where $d<m$ and $d>m$ denote over-complete and under-complete cases, respectively. The new feature embedding is then decoded back to an estimate ($\hat{X}$) as close as the original input via a decoder network. The encoder and decoder have a set of trainable parameters  $\theta\in\{ W_{\theta}, b_{\theta}\}$ and $\Phi\in\{ W_{\Phi}, b_{\Phi}\}$, respectively. The encoder (Z) and decoder ($\hat{X}$) outputs can be formulated as below.
\begin{eqnarray}
Z = f (\theta, X)\\
\hat {X} = g (\Phi, f (\theta, X))
\end{eqnarray}
The self-supervised objective function of autoencoders updates $\theta$ and $\Phi$ to minimize the sum of squared differences between input data (X) and the decoded output ($\hat {X}$) across all N training samples as below.
\begin{equation}
\xi  = \underset{\theta, \Phi}{\operatorname{argmin}} \sum^{N}_{i=1} ||  X_i - \hat{X_i} ||_2^2.
 \label{equation-LLE2}
\end{equation}

The gradient of Equation (3) with respect to $\theta$ is:
\begin{equation}
\frac {\delta \xi}{\delta \theta} = - 2 \sum^{N}_{i=1}(X_i - \hat{X_i}) \frac {\delta g (\Phi, Z_i)}{\delta\theta}.
\end{equation}
Here, f(.) and g(.) denote sigmoid activation functions for the encoder and decoder networks, respectively.  
\begin{eqnarray}
Z = \sigma (W_{\theta}X + b_{\theta}) \\
g (\Phi, Z) = \sigma (W_{\Phi}Z + b_{\Phi}) 
\end{eqnarray}
The derivative of a sigmoid function is 
\begin{equation}
\sigma' = \sigma (1 - \sigma). 
\end{equation}
Applying this derivative solution to Equations 5 and 6, we get
\begin{eqnarray}
\frac{\delta Z}{\delta\theta} &=& Z (1- Z) X,\\
 \frac {\delta g}{\delta\theta} &=& g (1- g)(W_{\Phi}\frac{\delta Z}{\delta\theta} + Z).
\end{eqnarray}
The trainable parameter $\theta$ is updated using a learning rate $\eta$ as follows.
\begin{eqnarray}
 \theta' = \theta + \eta \frac {\delta \xi}{\delta \theta}.
\end{eqnarray}
The error gradient can be summarized as follows. 
\begin{eqnarray}
 \frac {\delta \xi}{\delta \theta} = - 2\sum^{N}_{i=1}(X_i - \hat{X_i})\hat{X_i} (1- \hat{X_i})(W_{\Phi}\frac{\delta Z_i}{\delta\theta} + Z_i)
\end{eqnarray}

Therefore, an autoencoder does not involve data labels for weight updates. It can learn directly from the unlabeled data to preserve $d$-dimensional input data in an $m$-dimensional space. In the absence of target labels, autoencoders  yield a non-linear dimensionality reduction and a suitable initialization of the trainable parameters (pretraining) for subsequent classification tasks.

\section {Methodology}

This section discusses the proposed weight perturbation method, three deep neural networks used for experimentation, and the experimental steps and model evaluation procedures.

\begin{figure*}[t]
	\includegraphics [width=12cm]{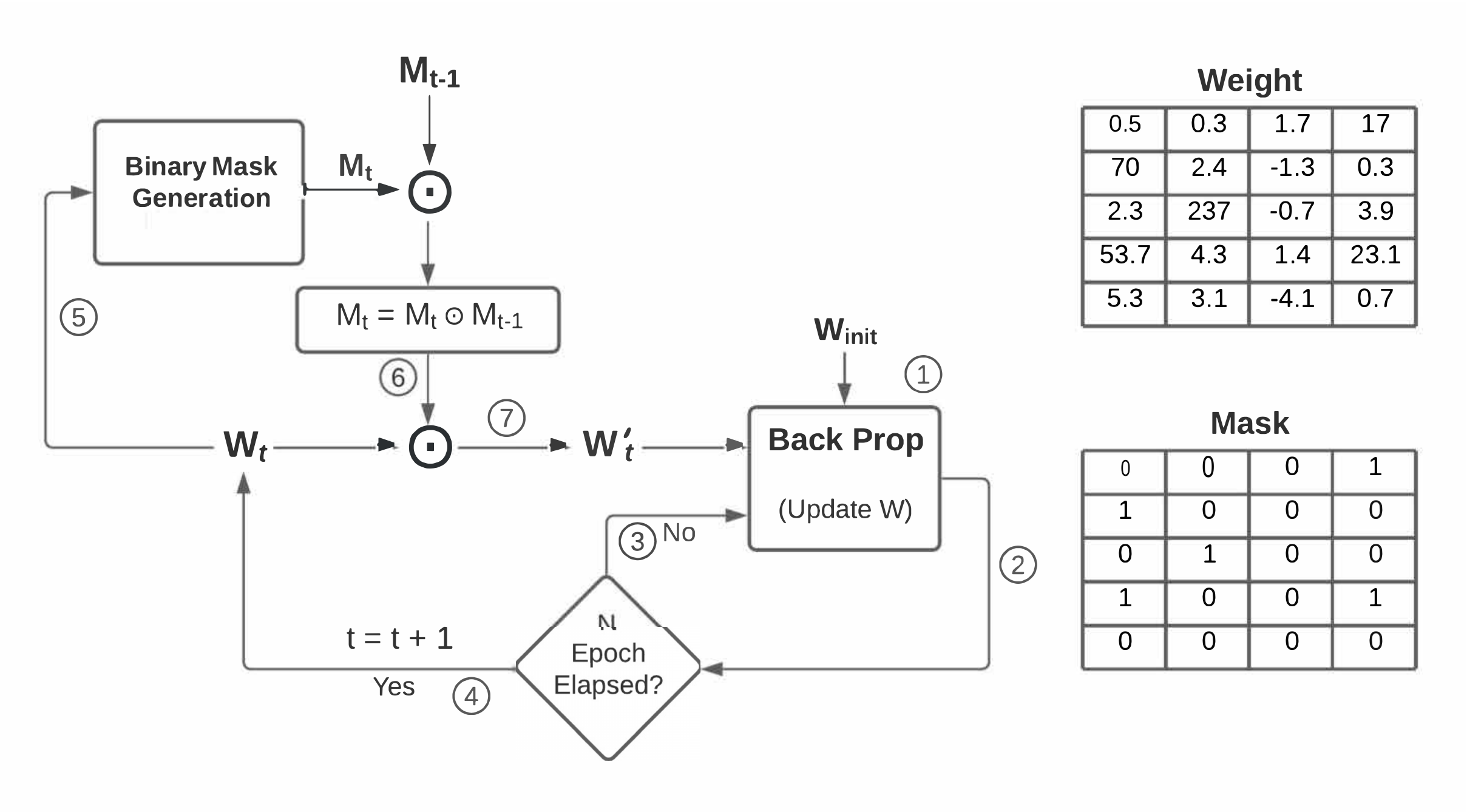}
	\vspace{-10pt}
	\caption{Sequential steps demonstrating periodic weight perturbation at N-epoch intervals. The mask contains zero in positions where the weights are to be pruned. After N epochs of training, a new mask ($M_t$) is generated (5) from the regrown weight values and then cumulatively included in the previous mask ($M_{t-1}$) by an element-wise product (6). The updated mask is used to prune the weights before the next back propagation (7).}
	\label{figdetailmethod}
\end{figure*}

\begin{figure*}[t]
\vspace{0pt}
	\vspace{-0pt}
	\includegraphics [width=8cm]{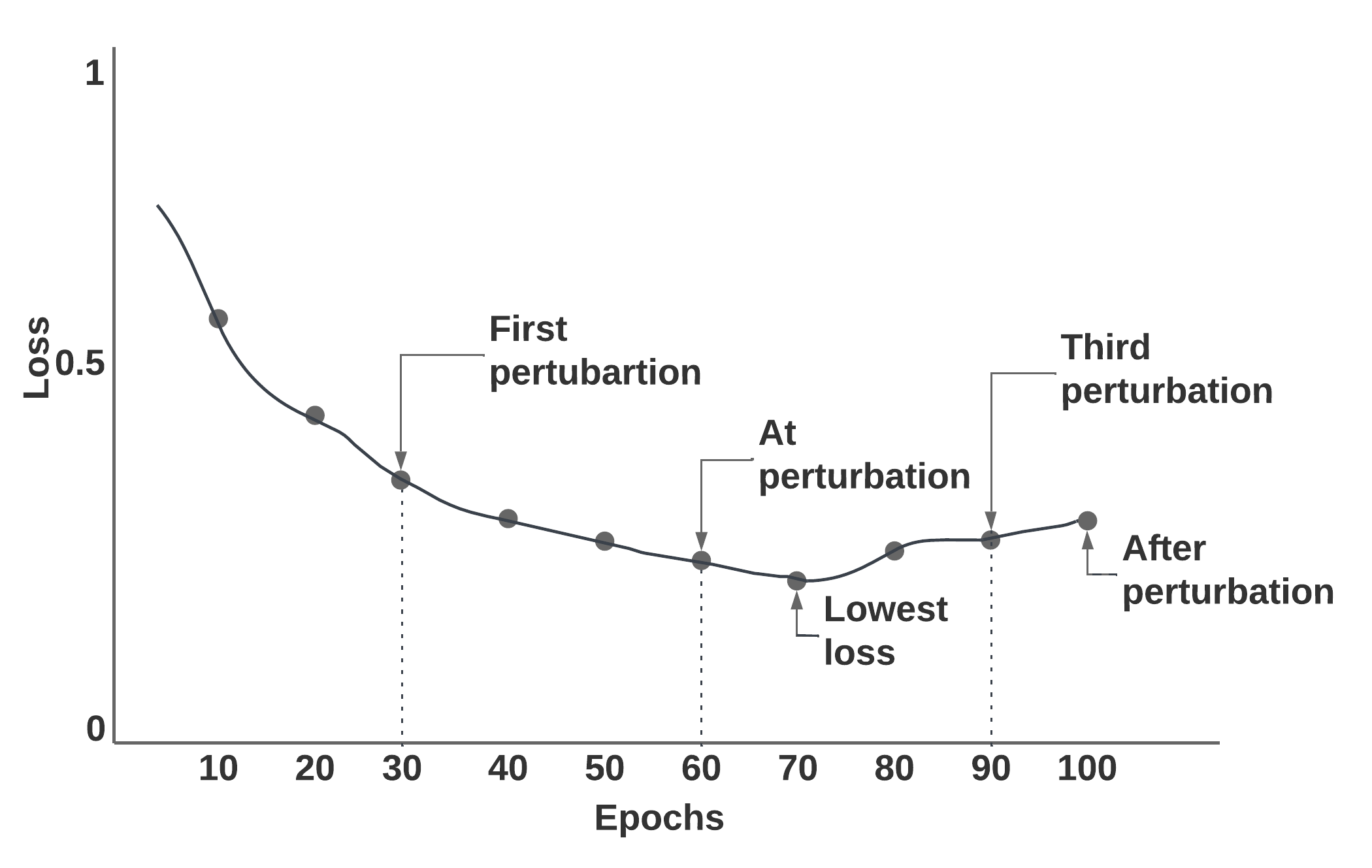}
	\vspace{-10pt}
	\caption{Illustration of weight perturbation schedule on an autoencoder training loss curve.}
	\label{fig:schedule}
\end{figure*}

\subsection {Periodic perturbation of weights}

We propose a weight pruning-and-regrowing routine during model training and term it weight perturbation. There are several differences between our method and those that exist in the literature~\citep{He_ACM_2018, Bellec_ICLR_2018, Mocanu_nature_2018}. First, our weight perturbation routine performs periodic pruning of weights at N$>$10 epoch intervals instead of every epoch to allow sufficient time to regrow some of the pruned weights. Second, the pruned weights are initialized by zero for regrowing instead of random initialization. Third, weights are perturbed only several times over the entire period of training instead of continuous pruning-and-regrowing. Last but not least, our weight perturbation method is proposed for feed forward deep neural networks learning tabular data rather than image filters of convolutional neural networks.

Let $W$ be the trained weight matrix updated via back propagation for $N$ epochs. After $N$ epochs, we obtain a binary mask for the first time (t=1) by selecting positions in the weight matrices based on the $\tau$\% threshold set on weight magnitudes as below. 
\[
    M_t= 
\begin{cases}
    0, & \text{if } min(W)*\tau < W < max(W)*\tau\\           
    1  & \text{otherwise}
\end{cases}
\]
The mask matrix contains 'zero' in positions where the weight values are less than $\tau$\% of the maximum weight value and greater than $\tau$\% of the minimum weight value. Zero entries in the mask are used to prune the corresponding weight values using the Hadamard product as W' = W$\odot$ M. For example, if the maximum and minimum weight values are +10 and -10, respectively, a $\tau=$5\% range is defined by $\pm$0.5. Any small weight values within the range of $\pm$0.5 are pruned using the binary mask. In the second perturbation time (t=2), following another N epochs of training and regrowing of the previously pruned weights, a new weight mask is obtained $M_{t}$ based on the new range of values defined by $\tau$\%. A Hadamard product between the previous mask ($M_{t-1}$) and the newly obtained mask ($M_{t}$) is obtained as $M_{t}$ = $M_{t} \odot$ $M_{t-1}$. Therefore, the mask matrix cumulatively stores the pruning positions at each N-epoch interval of training. This periodic masking (pruning) and regrowing procedure iteratively continues as shown in Figure~\ref{figdetailmethod}. The percentage of weights with zero values at a given time point of training is used to measure model compression or sparsity.

\begin{figure*}[t]
\subfigure[Basic DNN] {	\includegraphics [width=0.2\textwidth]{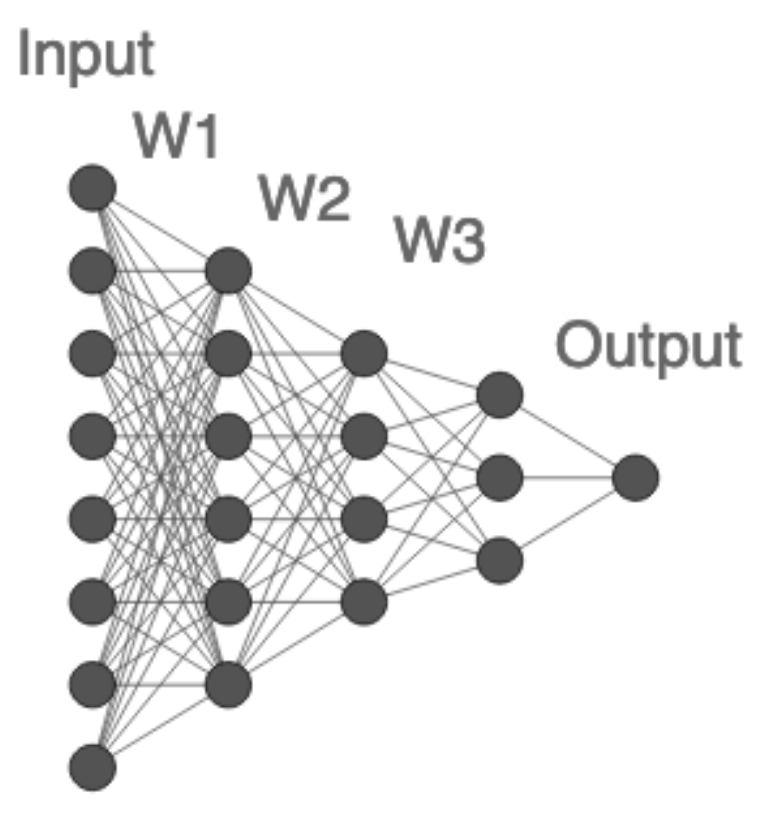}}
\subfigure[Basic Deep Autoencoder] {	\includegraphics [width=0.4\textwidth]{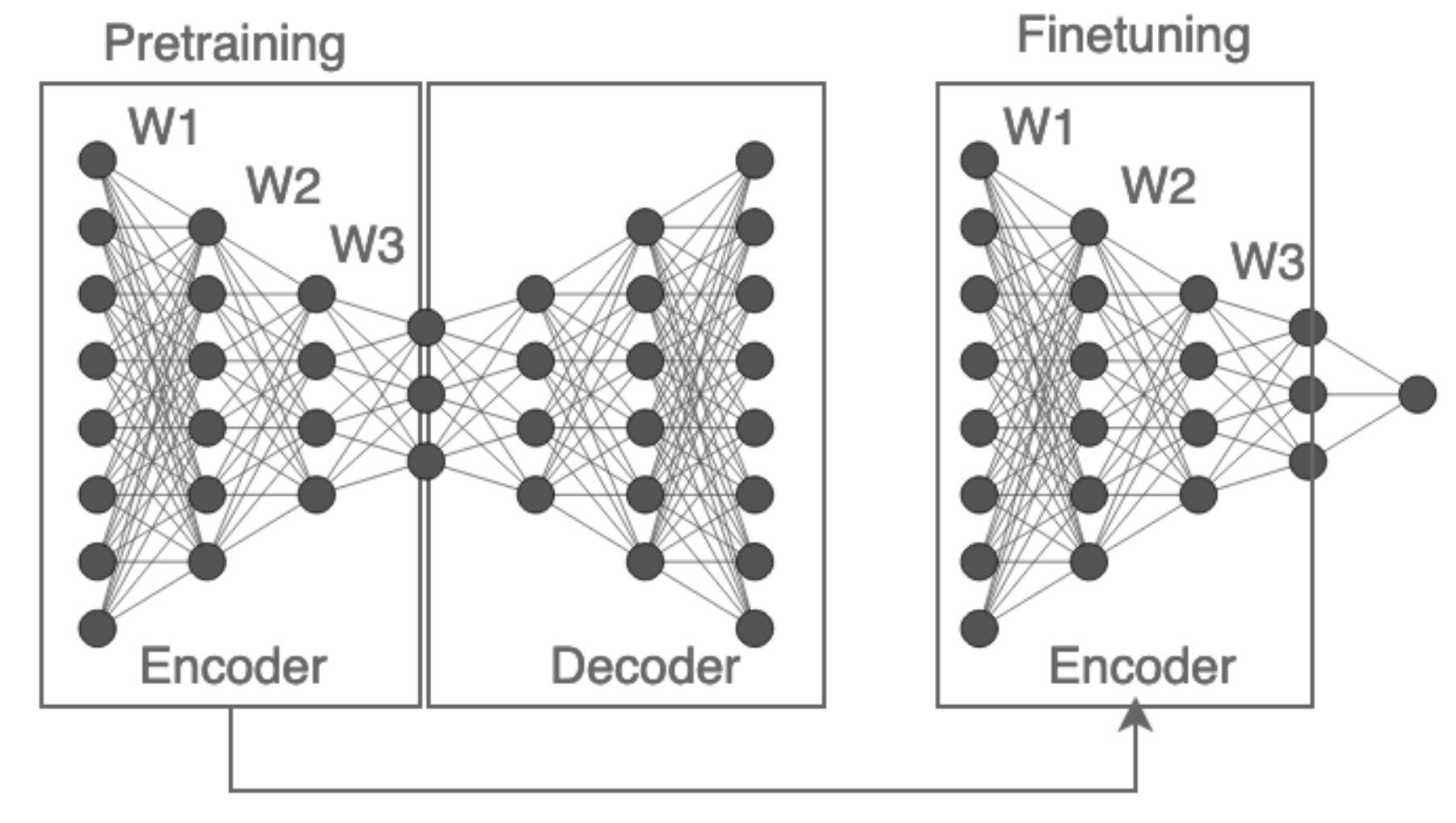}}
\subfigure[Stacked Deep Autoencoder]{\includegraphics [width=0.3\textwidth]{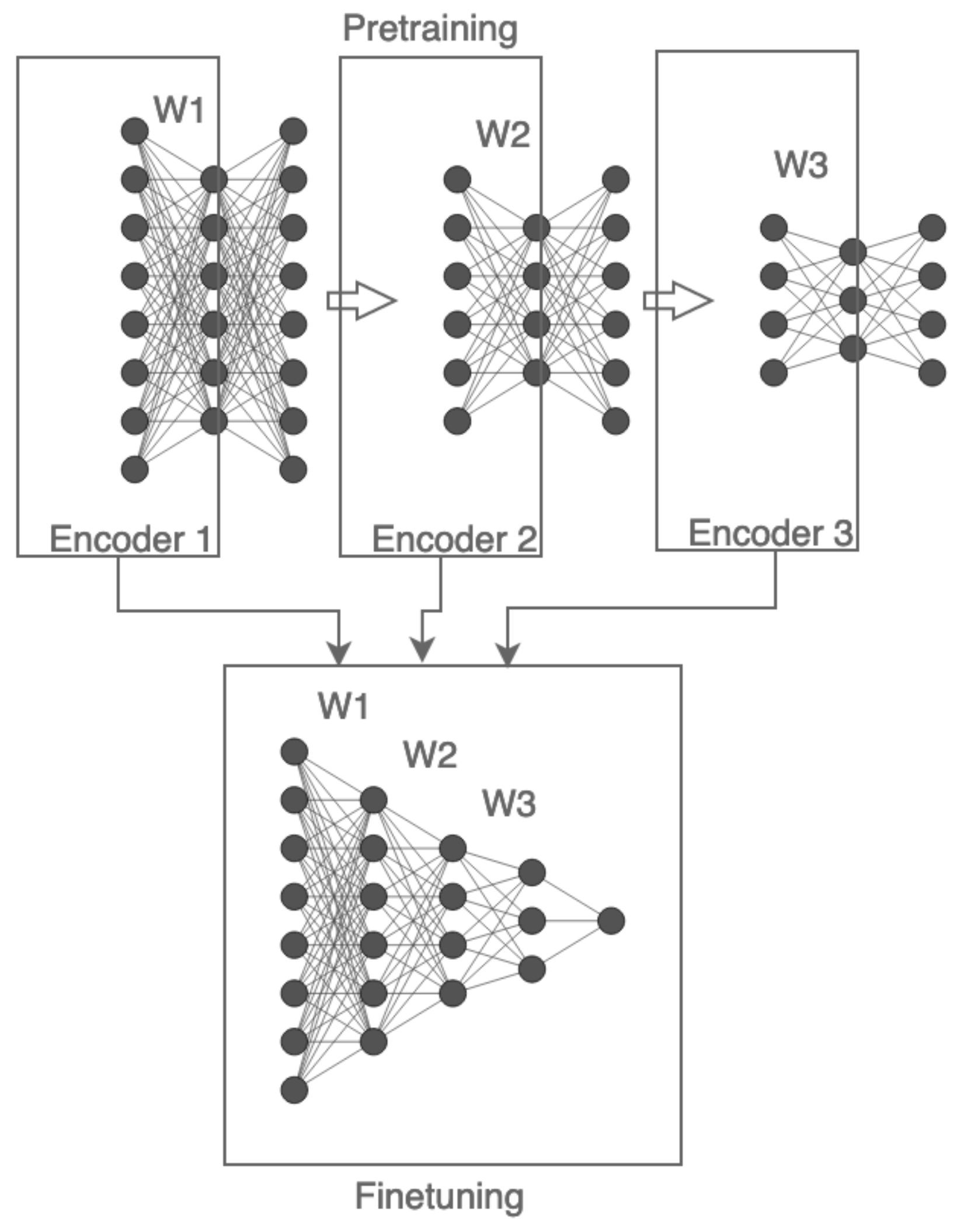}}
	\vspace{-10pt}
	\caption{Three feed forward deep neural network architectures used in this study.}
	\label{fig:deepmodel}
\end{figure*}

\subsection{Weight perturbation schedule and setting}

The best weight perturbation schedule is selected from our preliminary work~\citep{IJCNN}. Our
preliminary work on a single layer autoencoder reveals that periodically perturbing the weights of low magnitude several times can improve the data reconstruction loss. In this paper, we study the effect of proposed perturbation routine on pretraining a three-layer autoencoder (deep). This paper also sheds light on how such weight perturbations perform in transfer learning, especially when the pretrained autoencoder model is fine-tuned for a downstream classification task with tabular data. We extensively analyze the transfer learning performance of pretrained models obtained at different perturbation points, as shown in Figure~\ref{fig:schedule}. A pretrained encoder is taken from one of these time points for subsequent finetuning in a downstream classification task. A general notion is that the lowest point in the reconstruction loss is the best model to be chosen for downstream learning tasks.

\subsection {Deep model training}

We examine the proposed weight perturbation routine on three deep model: 1) a basic fully connected deep neural network (DNN), 2) a basic deep autoencoder (basic DAE), and 3) a stacked deep autoencoder (SDAE)~\citep{Zhou_2019}. Figure~\ref{fig:deepmodel} illustrates the three deep models, their architectures, and training mechanism. All three models have three hidden layers to comply with the definition of deep learning. The stacked autoencoder is formed by stacking three separately pre-trained autoencoders. In a stacked autoencoder, the first autoencoder hidden layer is trained using input data. The latent response of the first autoencoder is used to train the second autoencoder, and so on. The DNN model is trained from scratch and then tested on a set apart data set. Basic DAE and SDAE are first pretrained using the training data fold without any data labels. The pretrained models are then finetuned with the same training data fold with labels. Finally, the finetuned model is tested on a set aside test data fold.

\subsection{Experimental cases}

The proposed weight perturbation scheme is evaluated in five experimental cases: 1) no perturbation (baseline), 2) pretrained model at the lowest reconstruction loss with perturbation (lowest loss), 3) pretrained model at perturbation (right after a perturbation or pruning), 4) after perturbation (regrowing the weights after multiple perturbation points), and 5) only 20\% dropout. We investigate if the proposed perturbation method (case 2-4) improves feature learning over the baseline model with no perturbation (case 1) or dropout only (case 5). The performances of these experimental cases are evaluated in downstream classification tasks using three deep neural network models.

\subsection{Model evaluation}

We evaluate the experimental cases on three feed forward neural networks using six domain tabular data sets. The models are primarily analyzed in three steps: 1) pretraining with a training data set without labels, 2) finetuning with the training data set and corresponding labels, and 3) testing on a separate test data set. The train-and-test procedure is performed using a five-fold cross validation scheme where four folds are used to train (pre-train and fine-tune) the model, and the remaining fold is used to obtain the test accuracy on the finetuned model. The average F1-score across five test data folds is used as the model performance metric. The F1-score accounts for false positives and false negatives along with true positives, which is a robust metric for reporting classification performance.

\section{Results}

All experiments are implemented and executed on a Dell Precision 5820 workstation running Ubuntu 20.04 with 64GB RAM and an Nvidia GeForce RTX 3080 GPU with 10GB memory. The algorithms and model evaluation steps are implemented in Python. The deep learning models are developed using PyTorch. The findings of the experiments are discussed below.

\subsection{Tabular data sets} Six tabular data sets are obtained from the University of California at Irvine (UCI) machine learning data repository to train and test the models with or without the perturbation routine~\citep{Dua:2019}. Table~\ref{dataset_table} shows the summary statistics of the tabular data sets used in this paper. Four out of five data sets have binary classification labels. The gene data set has multiclass labels (five). The Madelon data set is synthetically generated for the NIPS 2003 feature selection challenge. For the malware data set, the class labels are two malware types: Virus Total (VT) and VxHeaven. The gisette data set contains handwritten characters '4' and '9'. The pixels of these images are vectorized and converted to tabular format. We select tabular data sets with relatively higher dimensionality ($>$500) than what is common in the literature. We assume that the sequential layers of deep neural networks are appropriate for learning features by reducing data dimensionality at each layer. No data set contains missing values, nominal variables, and ordinal variables. We encode the categorical class labels to numerical labels using label encoding for all data sets.

\begin{table*}[t]
\begin{tabular}{lcccc}
\toprule
Data set           & Sample size & Dimension & Classification & Domain                      \\ \midrule
Arcene             & 200         & 10001      & Binary              & Diagnostic                \\
Gene               & 801         & 16384      & Multiclass (5)      & Gene sequence             \\
Gisette            & 7000        & 5001       & Binary              & Handwritten digit         \\
Madelon            & 2600        & 501        & Binary              & Synthetic data                \\
Parkinson          & 756         & 754        & Binary              & Speech data \\
Malware & 5653        & 1087       & Binary              & Cyber-security                    \\ \bottomrule
\end{tabular}
\caption{A summary of the six tabular data sets used to evaluate the performance of proposed weight perturbation method.}
\label{dataset_table}
\end{table*}
\begin{figure*}[t]
\centering
\vspace{0pt}
\subfigure[Basic DNN and AE] {\includegraphics[trim=.6cm .6cm .6cm .6cm, width=0.45\textwidth] {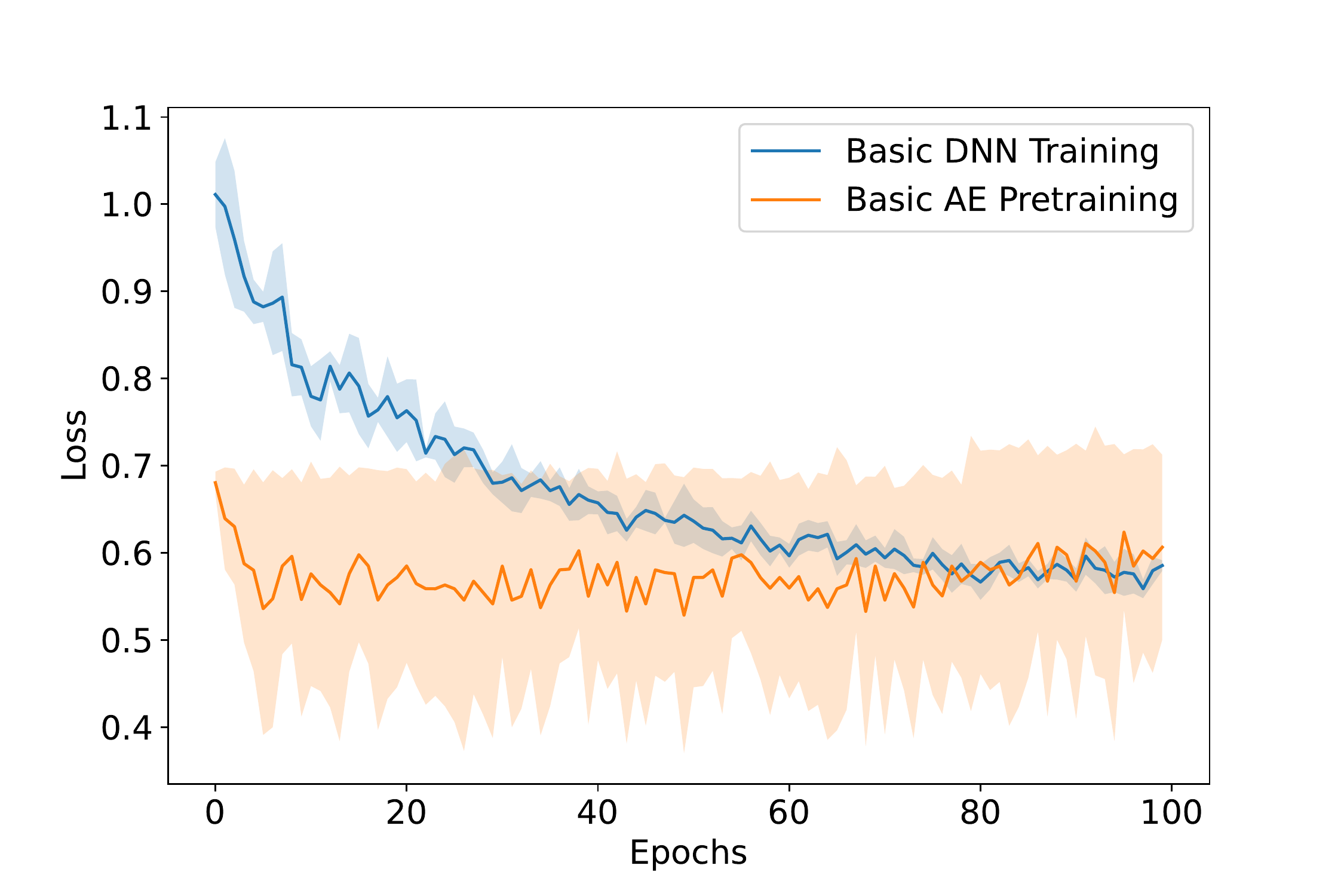}}
\subfigure[Perturbation versus Dropout] {\includegraphics[trim=.6cm .6cm .6cm .6cm, width=0.45\textwidth] {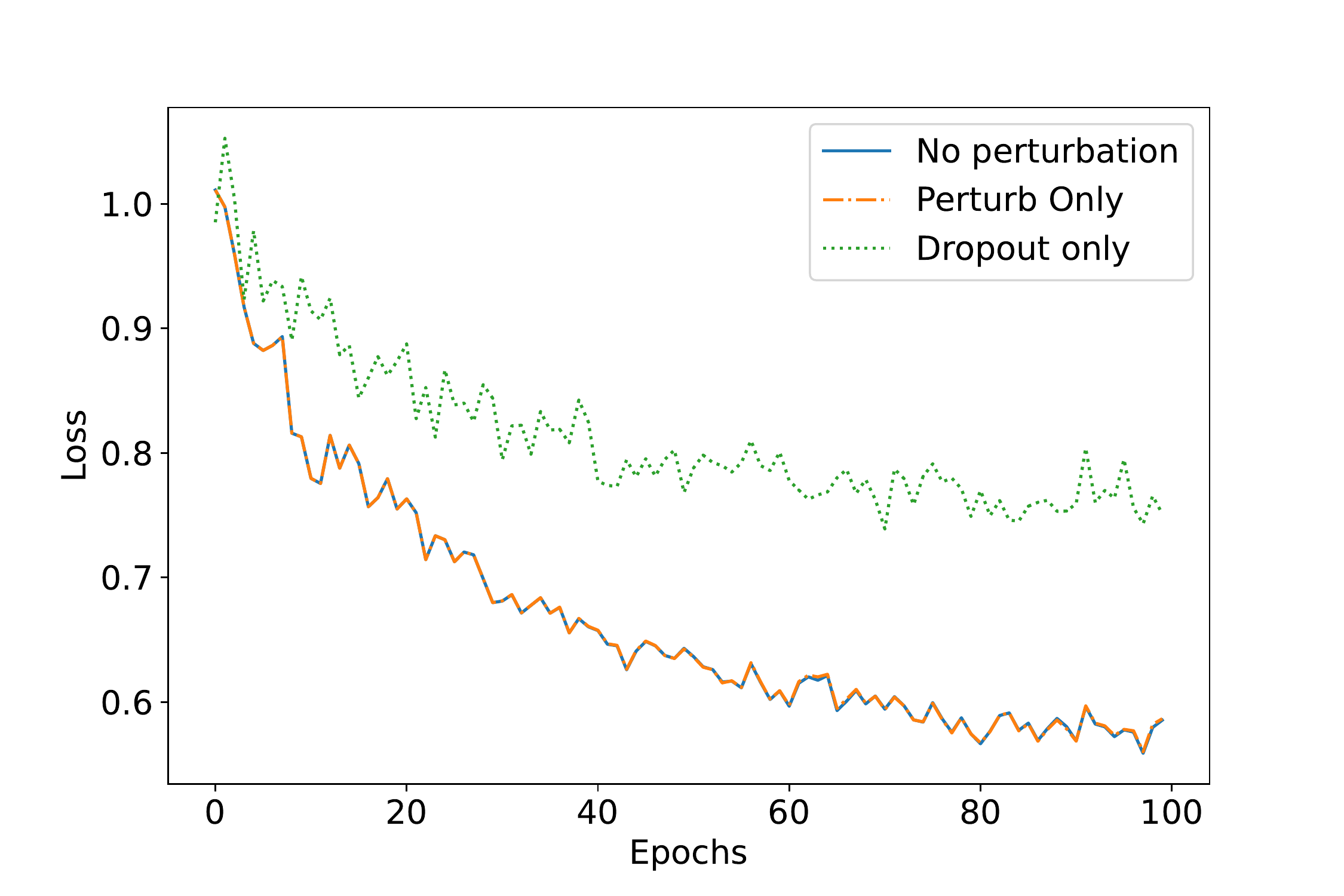}}\\
\vspace{-15pt}
\subfigure[Effect of $\tau$\%] {\includegraphics[trim=.6cm .6cm .6cm .6cm, width=0.45\textwidth] {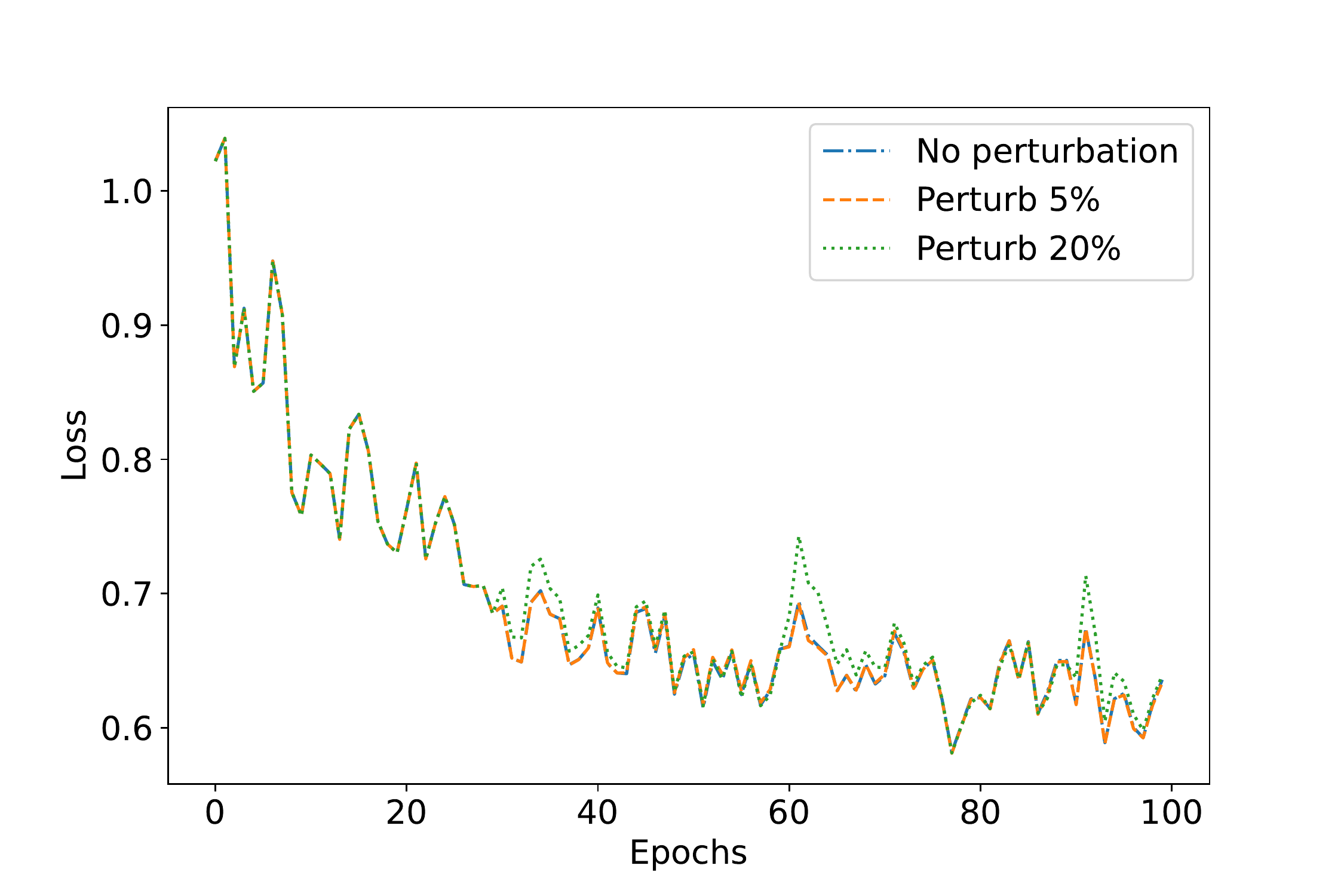}}
\subfigure[Deep Stacked AE] {\includegraphics[trim=.6cm .6cm .6cm .6cm, width=0.45\textwidth] {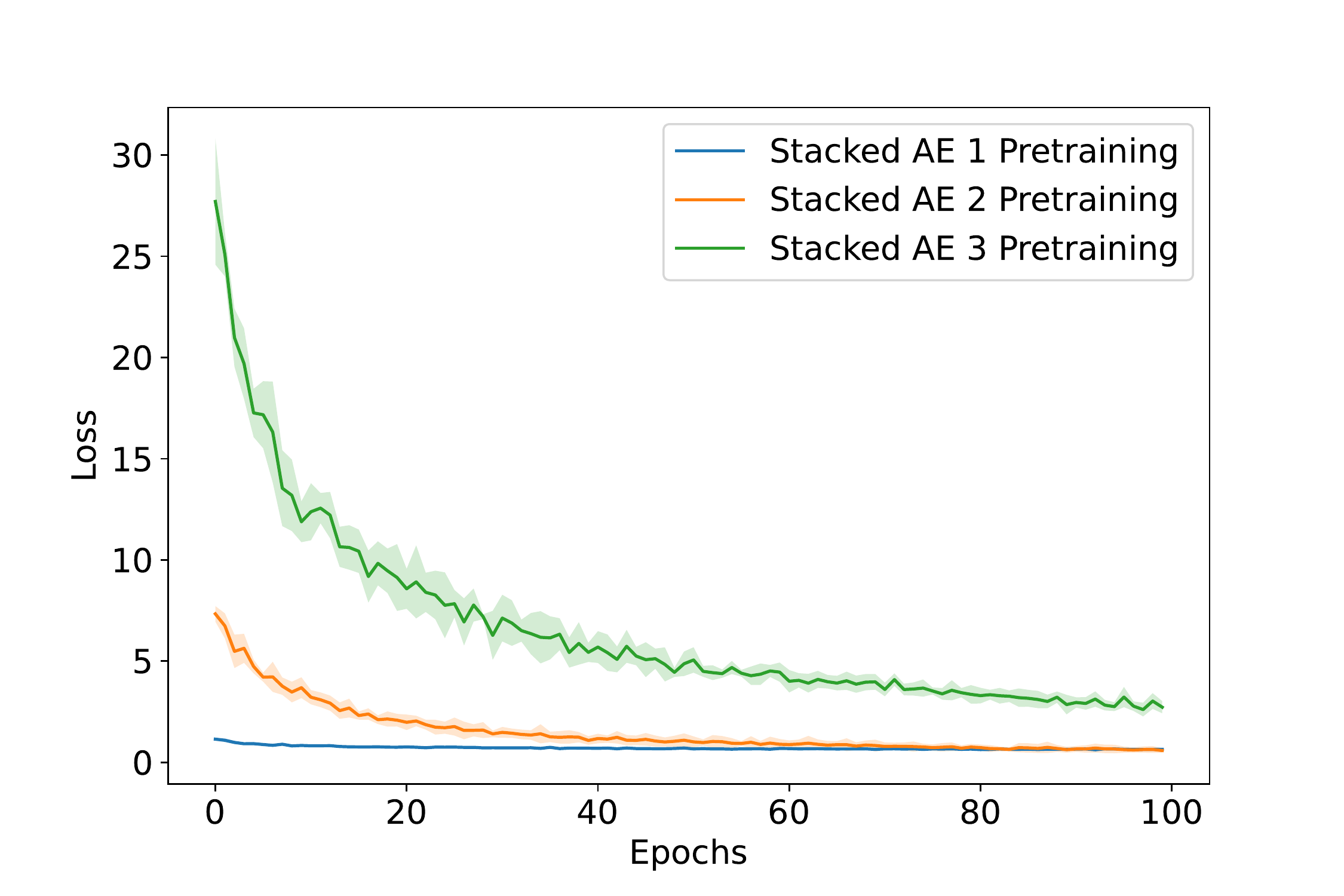}}
\vspace{-5pt}
\caption{Comparative illustrations of loss curves of different deep models and perturbation routines. Perturb 5\% donotes a $\tau$ value of 5\%. }
\label{figlosscurves}
\end{figure*}

\subsection{Data processing and deep model setup}
We standardize data using the mean and standard deviation of individual variables obtained from the training data folds. Therefore, no information about the test data is leaked to the training data folds. For all experimental cases and data sets, a common deep architecture is adopted using three hidden layers with 256, 128, and 64 neurons. The batch size is 64 for all data sets. For optimizing the weight update, Adam optimizer is used with a learning rate of $1e^{-3}$ without any weight decay. The exclusion of weight decay is because it introduces weight regularization, which may bias the experimental results. We use the rectified linear unit (reLu) activation function for all hidden layers. However, for the output classification layer, sigmoid and softmax are used in binary and multiclass classifications, respectively. In model pretraining, the loss function is mean squared error loss (MSEloss). The finetuning step uses binary cross-entropy loss (BCELoss) for all data sets except the gene data set. A cross-entropy loss is chosen for the Gene data set because it has five target classes. In the dropout experiment, the dropout rate is set to 20\%. Self-supervised pretraining and supervised fine-tuning are done separately for 100 epochs for DAE and SDAE. However, only supervised training is performed with the basic DNN model.

\subsection{Deep model pretraining}

Figure~\ref{figlosscurves} compares representative training loss curves for different models and experimental scenarios. The Figure~\ref{figlosscurves}(a) shows that the basic autoencoder pretraining loss (self-supervised) curve appears much lower than the basic DNN training loss (supervised). In Figure~\ref{figlosscurves}(b), the pretraining loss curve, due to dropout, stands above that of the baseline (no perturbation, no dropout) and perturbation cases. The effect of perturbation on the loss curve is not discernible may be because the threshold on weight magnitude $\tau$ is set to 5\% in this study. Figure~\ref{figlosscurves}(c) shows clearer effects of a larger $\tau$ (=20\%), where the loss curve spikes up compared to the curve without perturbation at the time of perturbation (Epochs 30, 60, and 90). Figure~\ref{figlosscurves}(d) shows pretraining loss curves for three separately trained autoencoders before stacking in deep stacked autoencoder. The shade represents the standard deviation of the loss across five folds. In a stacked autoencoder, the first encoding unit shows faster learning than the one used at upper encoding units. 

\begin{table*}[t]
\scalebox{.75}{
\begin{tabular}{@{}llccccc@{}}
\toprule
Data set                                &                      & Basic DNN    & Basic DAE                            & Stacked DAE                            & GBT                          & \% of weights pruned        \\ \midrule
                                        & Baseline             & 57.90 (11.7) & 80.48 (5.7)                         & 81.93 (6.9)                           &                              &                                \\
                                        & Lowest loss          & 57.90 (11.7) & 81.51 (4.4)                         & {\bf 84.98   (5.7)} &                              &                                \\
                                        & At perturbation      & 58.47 (12.0) & 81.47 (7.3)                         & 83.03 (4.6)                           &                              &                                \\
                                        & After perturbation & 57.90 (11.7) & 79.47 (5.2)                         & 83.95 (5.6)                           &                              &                                \\
 \multirow{-6}{*}{Arcene (200 x 10001)} & Dropout only         & 77.28 (7.2)  & 82.99 (6.9)                         & 84.50 (6.1)                          & \multirow{-6}{*}{82.8 (6.5)} & \multirow{-6}{*}{15.47 (3.7)}     \\
\midrule
                                        & Baseline             & 85.33 (16.9) & 99.62 (0.5)                         & 99.62 (0.5)                           &                              &                                \\
                                        & Lowest loss          & 85.33 (16.9) & 99.37 (0.6)                         & {\bf 99.75 (0.3)}                           &                              &                                \\
                                        & At perturbation      & 85.22 (16.8) & 99.62 (0.5)                         & 99.25 (0.7)                           &                              &                                \\
                                        & After perturbation & 85.22 (16.8) & 99.25 (0.6)                         & 99.50 (0.5)                           &                              &                                \\
\multirow{-5}{*}{Gene (801 x 16384)} & Dropout only         & 99.62 (0.5)  & 99.37 (0.6)                         & 99.63 (0.3)                           & \multirow{-5}{*}{99.2 (0.4)} & \multirow{-5}{*}{23.03 (9.7)}  \\
\midrule
                                        & Baseline             & 71.57 (31.9) & 98.49 (0.2)                         & 98.21 (0.1)                           &                              &                                \\
                                        & Lowest loss          & 71.57 (31.9) & 98.51 (0.2)                         & 98.30 (0.2)                           &                              &                                \\
                                        & At perturbation      & 71.54 (31.9) & {\bf 98.54 (0.3)} & 98.16 (0.1)                           &                              &                                \\
                                        & After perturbation & 71.62 (32.0) & 98.39 (0.2)                         & 98.17 (0.2)                           &                              &                                \\
\multirow{-5}{*}{Gisette (7000 x 5001)}             & Dropout  only         & 97.07 (0.8)  & 98.41 (0.2)                         & 98.43 (0.2)                         & \multirow{-5}{*}{97.6 (0.5)} & \multirow{-5}{*}{18.19 (0.7)}       \\
\midrule
                                        & Baseline             & 46.71 (9.0)  & 57.11 (1.5)                         & 55.67 (1.2)                           &                              &                                \\
                                        & Lowest loss          & 46.73 (9.1)  & 57.71 (2.4)                         & 55.89 (1.8)                           &                              &                                \\
                                        & At perturbation      & 46.24 (9.2)  & 58.12 (2.2)                         & 55.39 (1.6)                           &                              &                                \\
                                        & After perturbation & 46.01 (8.9)  & 57.78 (2.7)                         & 56.08 (2.2)                           &                              &                                \\
 \multirow{-5}{*}{Madelon (2600 x 501)} & Dropout only         & 50.49 (7.2)  & {\bf 64.97 (0.8)} & 56.90 (1.5)                           &                              \multirow{-5}{*}{81.6 (1.9)} & \multirow{-5}{*}{20.09 (0.6)} \\
\midrule
                                        & Baseline             & 57.56 (37.1) & 88.17 (1.6)                         & 88.13 (2.8)                           &                              &                                \\
                                        & Lowest loss          & 58.54 (37.9) & 87.89 (2.5)                         & 87.98 (2.0)                           &                              &                                \\
                                        & At perturbation      & 58.52 (37.9) & 87.39 (2.2)                         & 87.54 (3.1)                           &                              &                                \\
                                        & After perturbation & 57.26 (36.9) & 88.06 (1.8)                         & 87.78 (1.9)                           &                              &                                \\
 \multirow{-5}{*}{Parkinson's (756 x 754)}  & Dropout only         & 72.14 (29.0) & {\bf 88.98 (2.8)} & 88.01 (2.5)                          & \multirow{-5}{*}{87.4 (1.4)} & \multirow{-5}{*}{17.82 (0.7)} \\
\midrule
                                        & Baseline             & 68.19 (30.6) & 94.37 (0.5)                         & 94.30 (7.0)                           &                              &                                \\
                                        & Lowest loss          & 68.22 (30.6) & 94.36 (0.7)                         & {\bf 94.69 (1.0)}                           &                              &                                \\
                                        & At perturbation      & 68.12 (30.6) & 93.43 (1.3)                         & 94.67 (0.7)                           &                              &                                \\
                                        & After perturbation & 68.47 (30.8) & 94.34 (0.4)                         & 94.69 (0.3)   &                              &                                \\
\multirow{-5}{*}{Malware (5653 x 1087)}  & Dropout only         & 93.33 (0.7)  & 94.55 (0.6)                         & 94.32 (0.6)                          & \multirow{-5}{*}{96.6 (4.9)} & \multirow{-5}{*}{41.28 (1.9)}           \\
\midrule
\end{tabular}
}
\caption{Average F1-scores across five cross-validation folds comparing different model perturbation scenarios. Basic DNN = Deep Neural Network and DAE = Deep AutoEncoder. The  pretraining is performed 1) with no perturbation or dropout (Baseline), 2) at the lowest autoencoder reconstruction loss point during perturbation  (lowest loss), 3) at the second perturbation point (at perturbation), 4) at epoch 100: regrowing after the final perturbation at epoch 90 (after perturbation). GBT = gradient boosting tree classifier model.  The last column shows the percentage of total weights pruned by the best pretraining scenario.}
\label{bigtable}
\end{table*}

\subsection {Data set specific trends}

Each data set is unique in sample size, dimensionality, sparsity, and distribution. Table~\ref{bigtable} highlights data set specific F1-scores due to different model regularization schemes. For the acrene data set, the baseline model (basic DNN, basic AE, and stacked AE) performance is generally improved by weight perturbation. The dropout learning substantially improves the performance of basic DNN. However, the performance gain due to dropout learning is on par with weight perturbation when using the autoencoder models. The highest F1-score (84.98 (5.7)\%) is obtained using a weight perturbed stacked autoencoder at the lowest pretraining loss, whereas dropout learning yields (84.5 (6.1)\%). The stacked autoencoder, with the aid of weight perturbation or dropout learning, outperforms the Gradient Boosting Tree (GBT) classifier performance (F1-score: 82.8 (6.5)\%). Otherwise, traditional machine learning (GBT) outperforms all three deep learning models at baseline. For the gene sequence data set, the pretraining of baseline deep autoencoders yields a large improvement in accuracy over basic DNN at baseline. The performance of basic AE with dropout learning alone slightly drops from the baseline and then further improves in joint dropout and weight perturbation learning. The highest F1-score (99.75 (0.3)\%) is obtained using a weight perturbed stacked autoencoder at the lowest pretraining loss when compared to the dropout learning (F1-score 99.63 (0.3)\%). Except for the basic DNN, all autoencoder-based deep models outperform the GBT model (F1-score 99.2 (0.4)\%). For the Gisette data set, the basic DNN never outperforms traditional machine learning model (GBT) performance. The autoencoder-based models invariably outperform the GBT classifier. The highest F1-score is achieved by the basic autoencoder when taken at the time of perturbation (at perturbation, F1-score 98.54(0.3)\%) for finetuning. This best score slightly outperforms the one achieved by dropout learning (98.41(0.2)\%). The Madelon data set is unlike the other five data sets because it is synthetically generated under some statistical assumptions. The GBT classifier (F1-score 81.6(1.9)\%) substantially outperforms any other deep learning models, considering all perturbation and dropout schemes (64.97(0.8)\%). This is an example where deep learning fails miserably without leaving an alternative to traditional machine learning. For basic DNN and basic AE, the dropout learning significantly improves the accuracy over the baseline models. Any weight perturbed models with the lowest loss slightly outperform their baseline counterparts. However, the best F1-scores are achieved by dropout learning (F1-score 64.97 (0.8)\%) using a basic pretrained AE model. In the Parkinson's data set, the performance of weight perturbation slightly drops compared with the baseline autoencoder models. The same is observed for dropout learning with the stacked autoencoder model. The baseline autoencoders (F1-score 88.17(1.6)\%) outperform the GBT classifier (F1-score 87.4(1.4)\%). The best F1-score is achieved by dropout learning with a basic autoencoder (F1-score 88.98(2.8) \%). Like the Madelon data set, none of the deep learning scenarios outperform the GBT classifier (F1-score 96.6(4.9 \%) for the malware data set. The perturbation of weights improves F1-score with the stacked autoencoder model, whereas dropout improves all baseline model performances. 
The best performance is achieved after weight perturbing a stacked autoencoder (F1-score 94.69(0.3)\%), which still falls below the accuracy achieved by a traditional machine learning classifier.

\subsection {Model specific trends}
Table~\ref{bigtable} reveals that self-supervised pretraining of a deep neural network (autoencoders) significantly improves its downstream classification performance over a basic deep classifier model (basic DNN) without pretraining. This observation on baseline models is consistent across all domain data sets. However, the contribution of the proposed weight perturbation varies across the deep models. The pruning step in perturbation compresses the model, and the regrowing step recovers some of the pruned weights to improve the pretraining of model. The basic autoencoder and stacked autonencoder models equally share the best performance counts (three/three) across six data sets. However, the baseline basic DNN model fails to outperform the machine learning model (GBT) on all six datasets. In contrast, with the aid of weight perturbation or dropout, deep autoencoder models outperform GBT (traditional machine learning) on four out of six data sets.

\subsection {Model compression by weight perturbation}
\begin{figure*}[t]
\subfigure[Arcene] {	\includegraphics [width=0.32\textwidth]{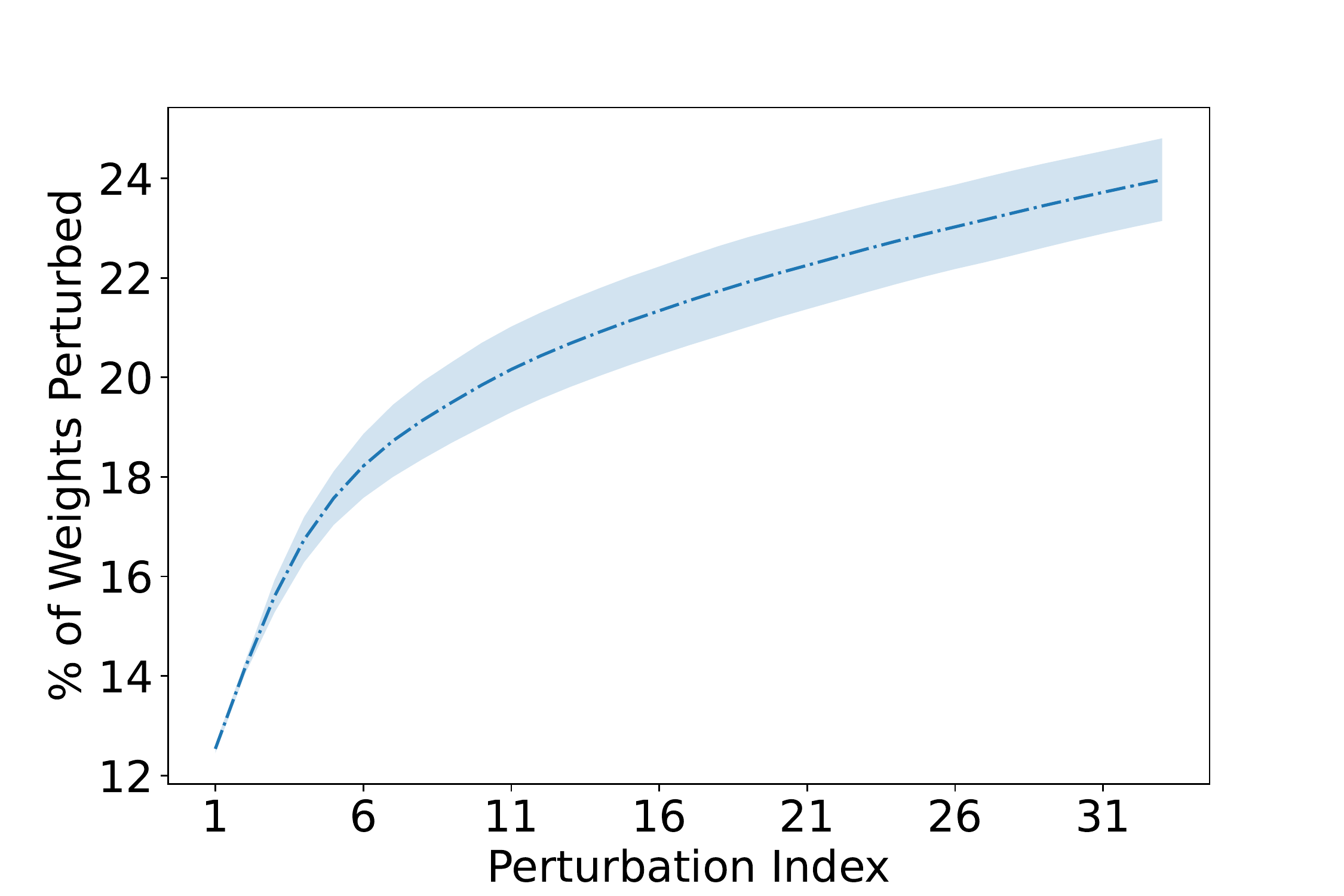}}
\subfigure[Gene ] {	\includegraphics [width=0.32\textwidth]{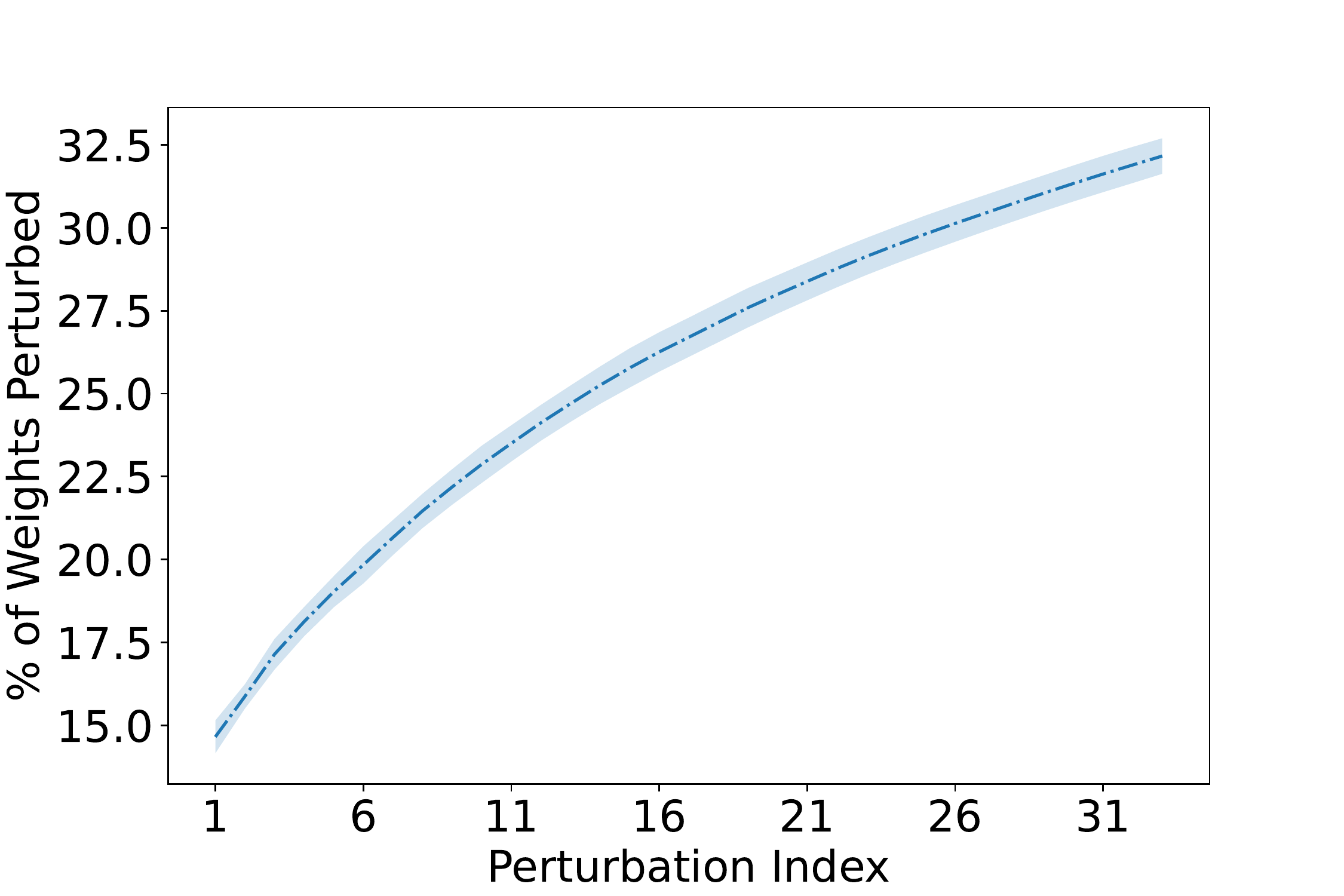}}
\subfigure[Gisette]{\includegraphics [width=0.32\textwidth]{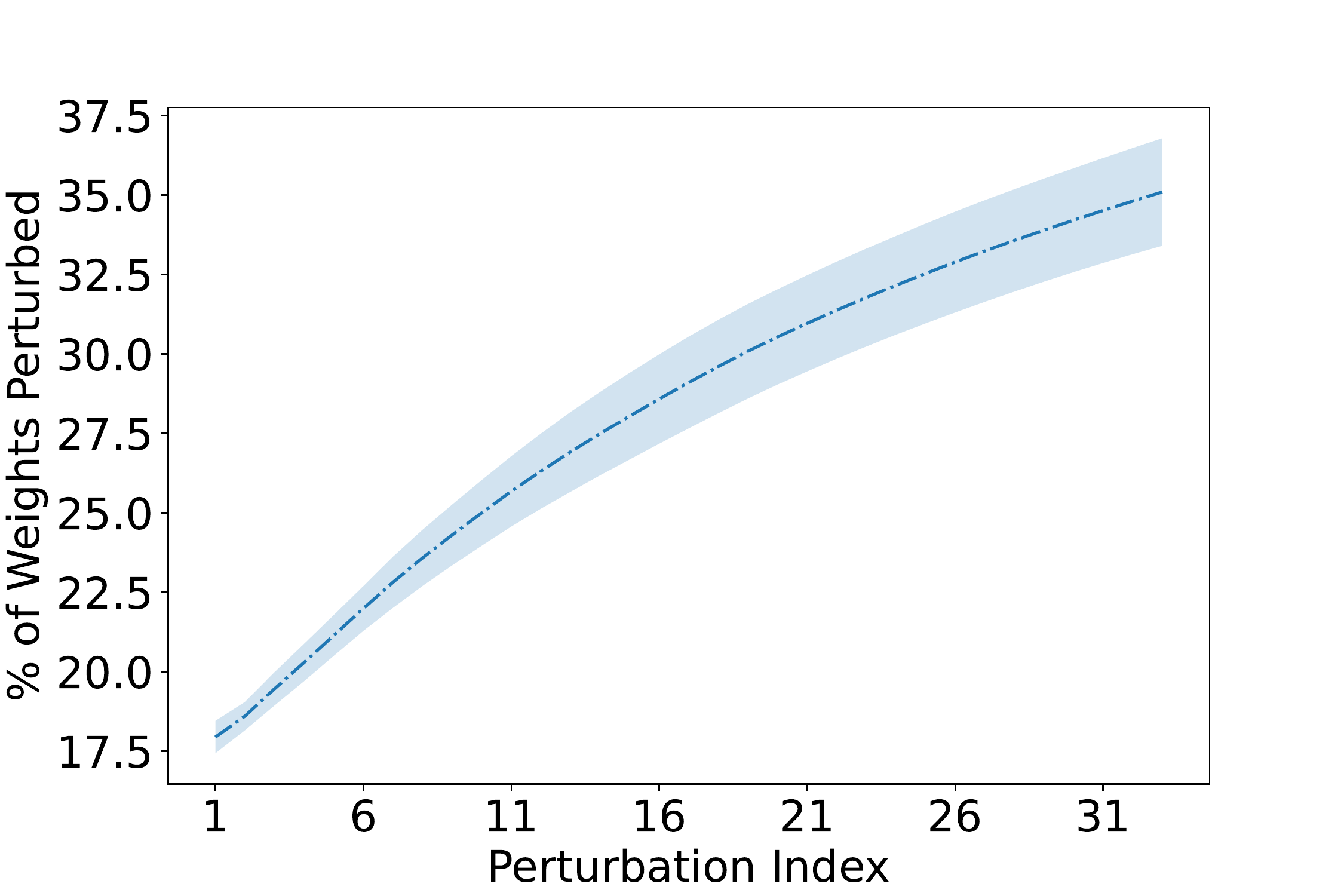}}
\subfigure[Madelon] {	\includegraphics [width=0.32\textwidth]{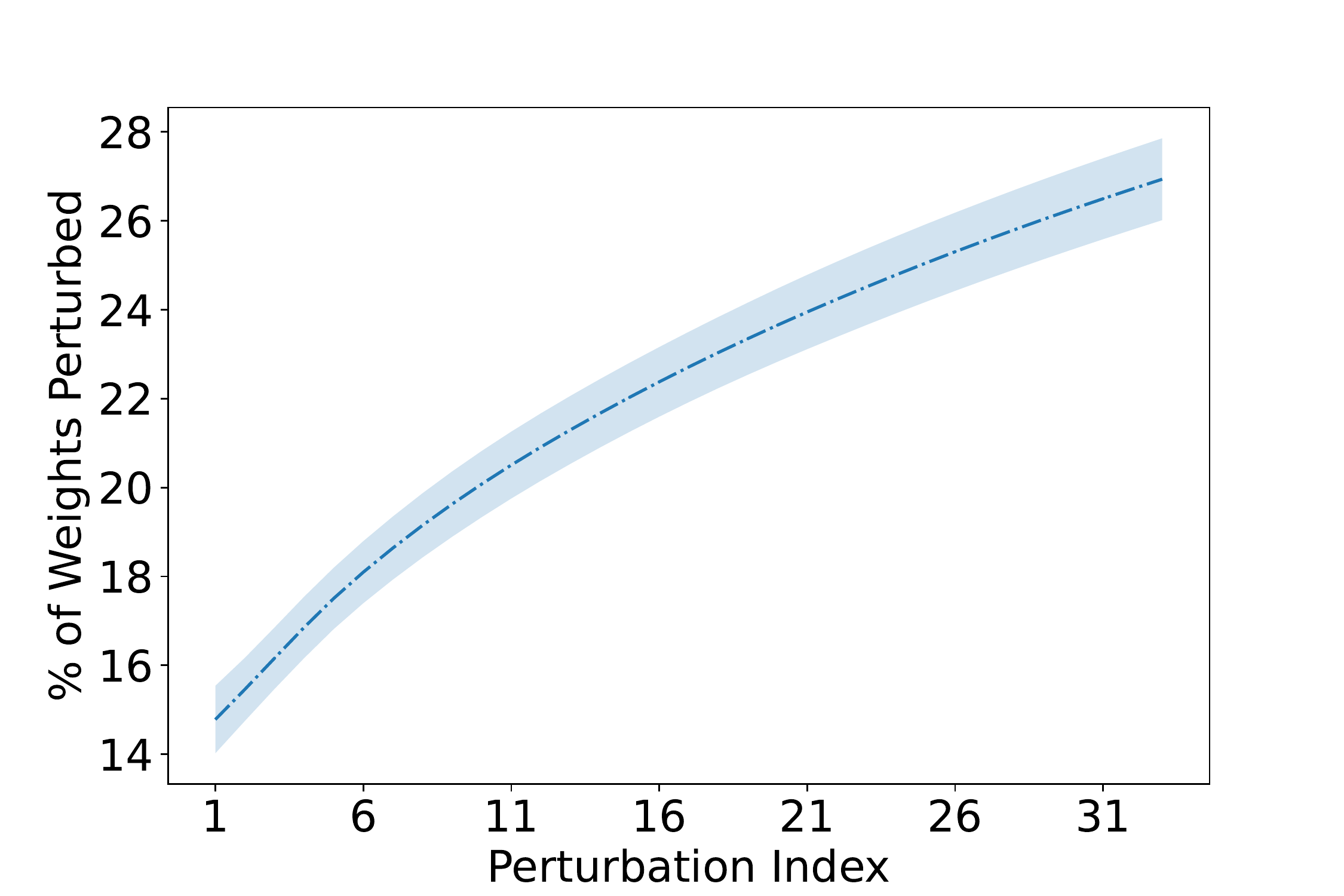}}
\subfigure[Malware] {	\includegraphics [width=0.32\textwidth]{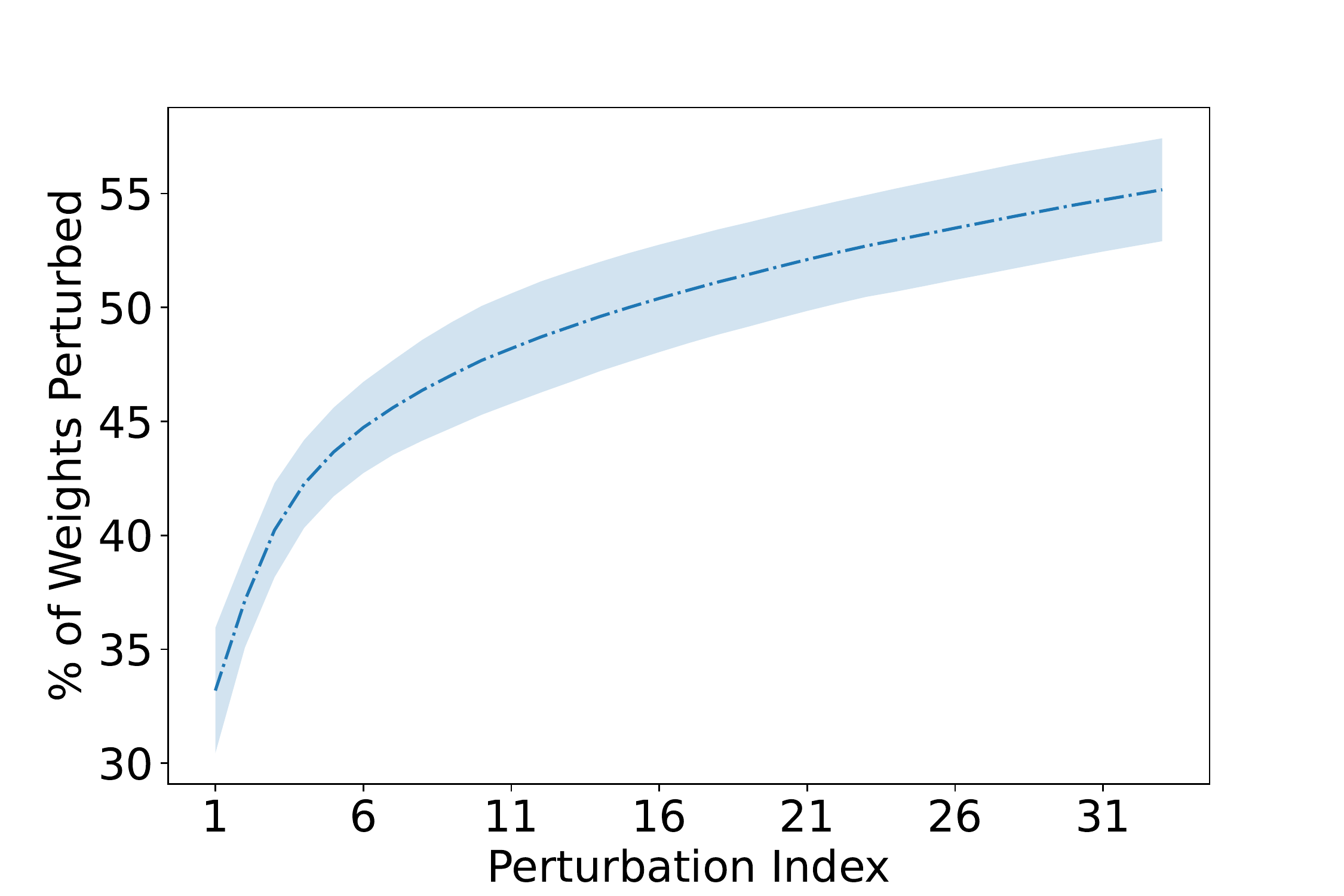}}
\subfigure[Parkinson's]{\includegraphics [width=0.32\textwidth]{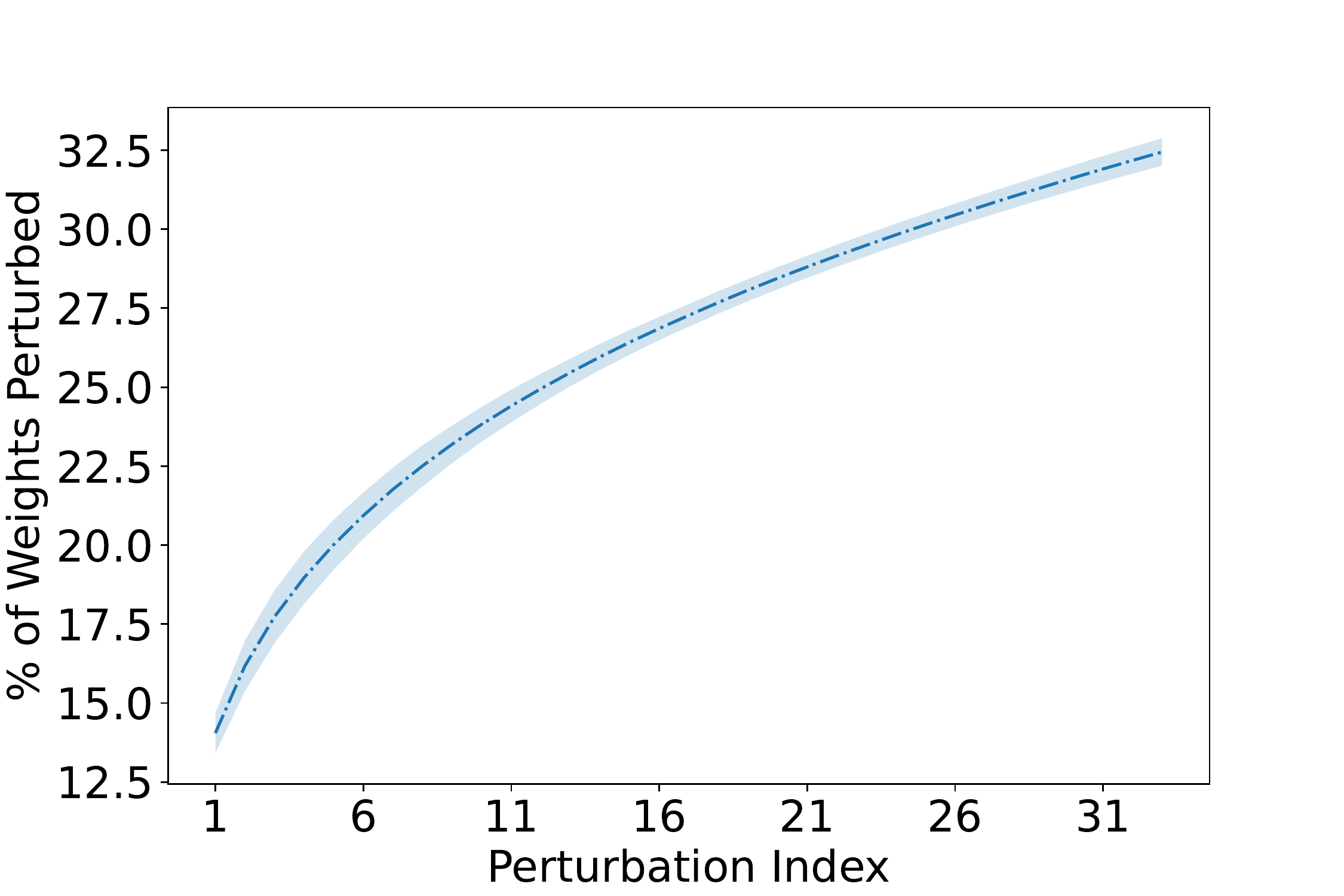}}
 	\vspace{-5pt}
\caption{Cumulative growth in the percentage of weights perturbed. The pretraining is continued for 1000 epochs to perturb the weights 33 times at 30-epoch intervals. }
\label{fig:1000_perturbation}
\end{figure*}

In addition to the performance gain in downstream classification, the pretrained model achieves a sparse weight representation. Table~\ref{bigtable} shows the percentage of weights pruned in the pretrained models that yield the best F1-scores for individual data sets after finetuning. The perturbation prunes 15\% to 40\% of the total weights in the pretrained model, resulting in significant model compression for storing and sharing such models to be used in future finetuning or supervised classification. Low standard deviations in the average pruning number suggest a stable selection of weights for pruning across the five data folds. The cumulative weight mask generation ($M_t$) via weight perturbation is observed by pretraining a deep model for 1000 epochs. Figure~\ref{fig:1000_perturbation} shows that after increasing with a larger slope, the percentage of weight pruned by the weight mask tends to slow down over subsequent training epochs.

\section {Discussion}

This paper proposes a method to periodically perturb weights of deep neural networks during training to simultaneously improve feature learning and model compression. The findings of the paper can be summarized as follows. First, the pretrained model at the lowest reconstruction loss point with weight perturbation improves downstream classification performance. Second, our proposed weight perturbation achieves 15\% to 40\% compression in pretrained models across six different tabular data sets. Third, instead of conceding a loss in accuracy, as seen in conventional weight pruning, our sparse pretrained deep model improves downstream classification performance. Fourth, the proposed weight perturbation method outperforms dropout learning in four out of six tabular data sets and is on par with dropout in the other two datasets. Therefore, weight perturbation may help alleviate some overfitting and introduce model compression. Fifth, a traditional machine learning model outperforms basic DNN models in tabular data classification. In contrast, pretrained autoencoders with weight perturbation can outperform a traditional machine learning model on similar data.  

\subsection {Effects of regrowing after pruning}
Conventional weight pruning methods permanently set some weight values to zero. In weight perturbation, the pruned weights are set to zero and allowed to regrow to recover some of the lost weights. The regrowth of weights after pruning can improve the learning trajectory of the pretrained model over time (1000 epochs)~\citep{IJCNN}. Without regrowing the pruned weights, pruning alone yields a negative effect by increasing the reconstruction loss. However, the effect of weight pruning with $\tau = 5\%$ is not evident in our loss curves. This observation may be due to our deeper networks with a large number of trainable parameters and a limited period of training (100 epochs) (Figure~\ref{figlosscurves} (b)). The regrowth of weights adjusts the loss curve, after retaining a subset of the sparse weights.

\subsection{Perturbation versus dropout and regularization}

Dropout is a random process of regularizing neurons during model training. It prevents the contribution of a set of randomly selected neurons to weight updates in the training phase. However, all neuronal responses are taken into account in the test phase. Therefore, the pretrained model obtained following dropout learning is not sparse. Despite the regrowing of weights, a significant portion of the weights remains zero, introducing a decent sparsity in the trained model. It is noteworthy that the F1-scores from perturbed models are compared with those obtained following L1 and L2 weight regularization during training. Weight regularization during the self-supervised pretraining phase yields worse F1-scores in subsequent classification tasks than dropout. It may be because weight regularization shrinks the pretrained weights, aiming to minimize the data reconstruction loss. The weight shrinkage may be too aggressive to favorably start and converge the pretrained model in supervised finetuning. For a fair comparison, dropout, weight perturbation, and weight regularization are performed during the pretraining phase only.     

\begin{figure*}[t]
\subfigure[Arcene] {	\includegraphics [width=0.32\textwidth]{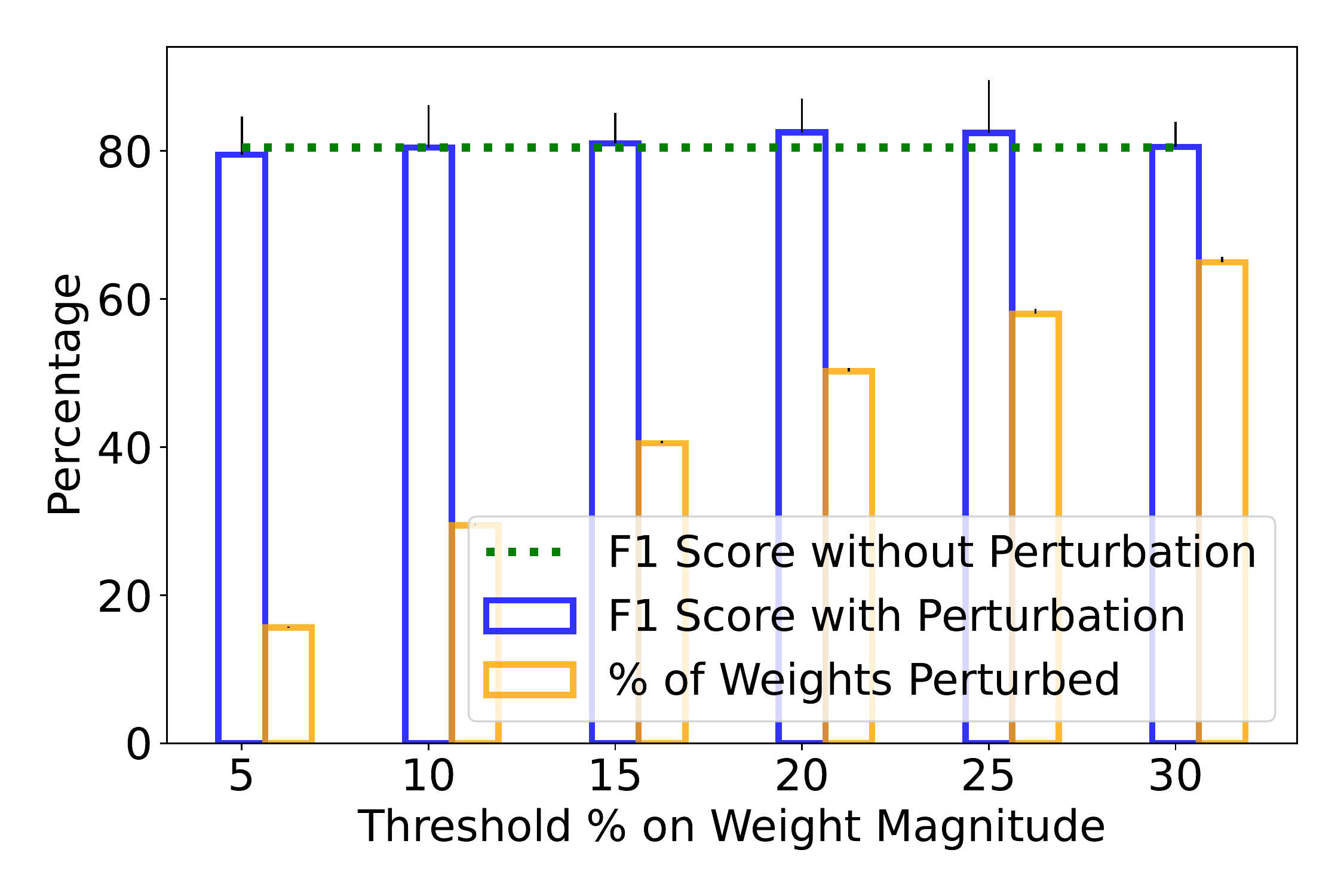}}
\subfigure[Gene ] {	\includegraphics [width=0.32\textwidth]{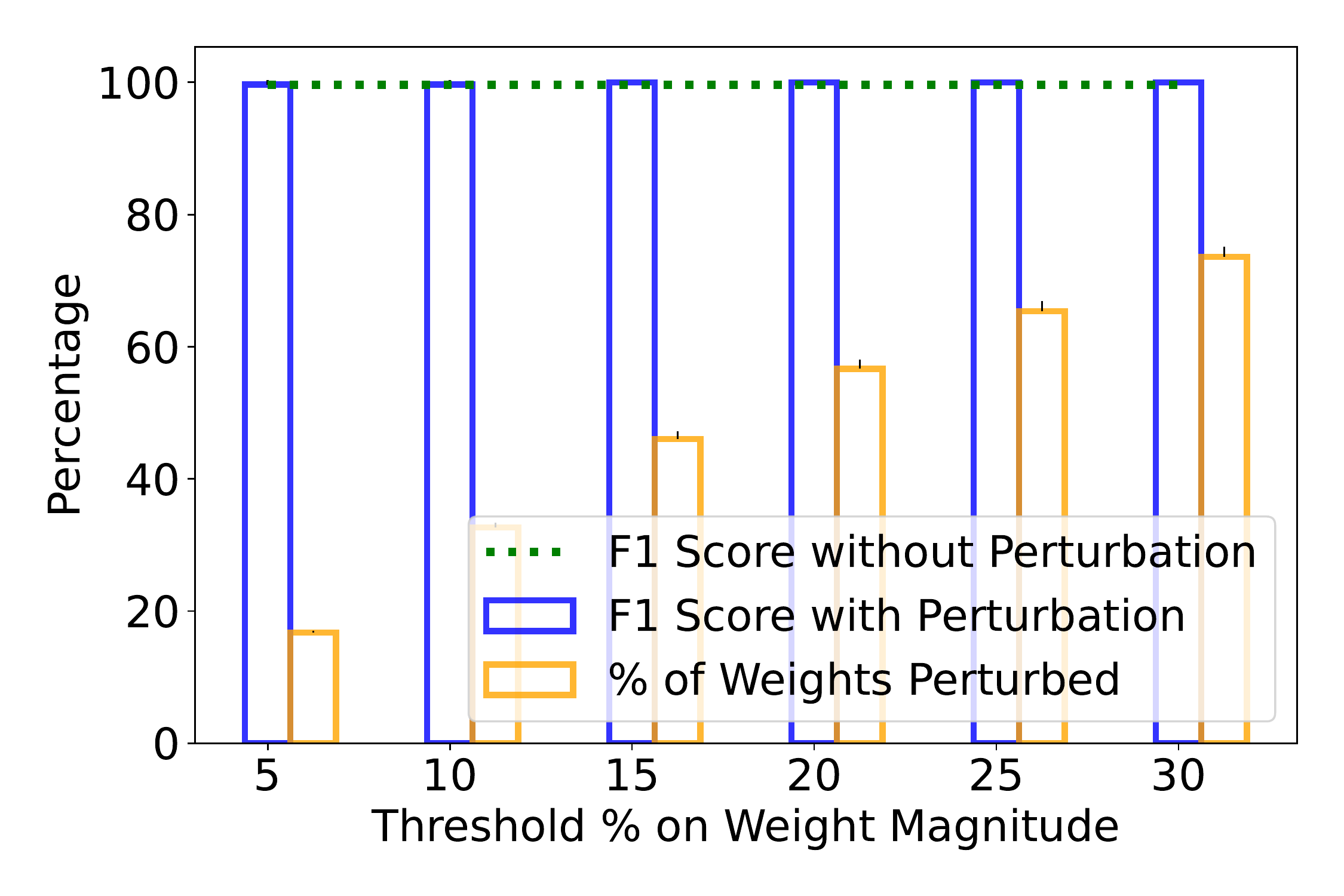}}
\subfigure[Gisette]{\includegraphics [width=0.32\textwidth]{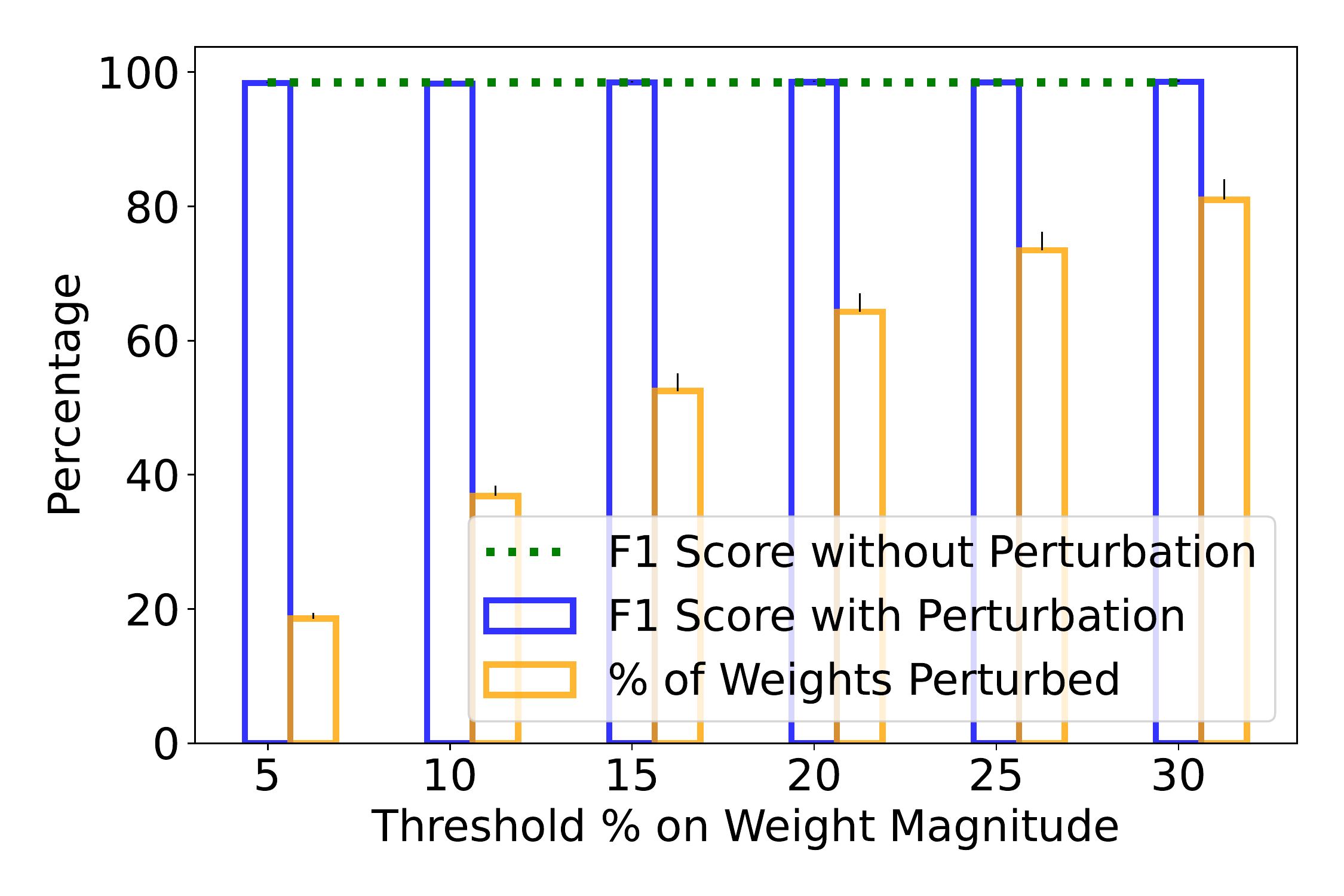}}
\subfigure[Madelon] {	\includegraphics [width=0.32\textwidth]{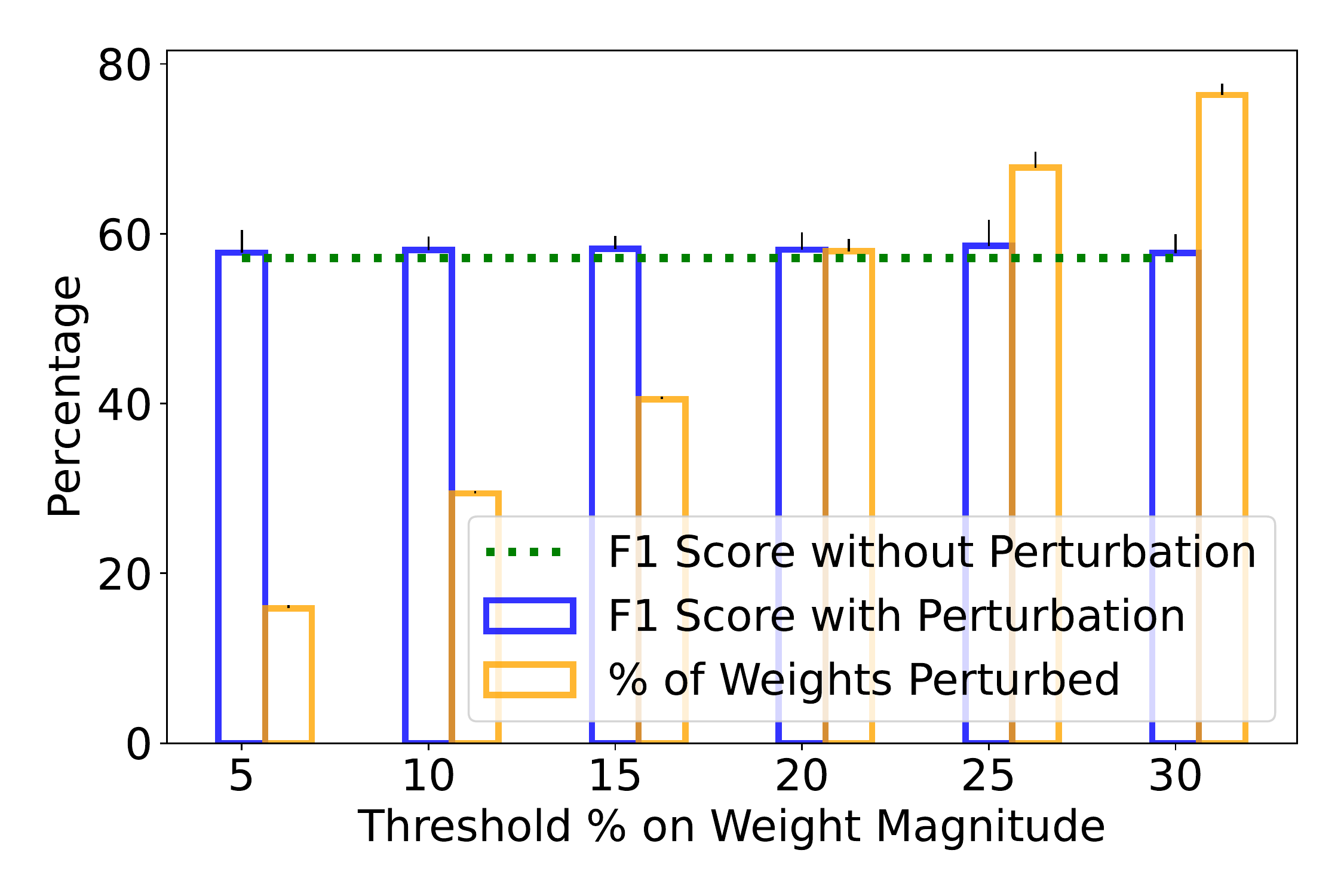}}
\subfigure[Malware] {	\includegraphics [width=0.32\textwidth]{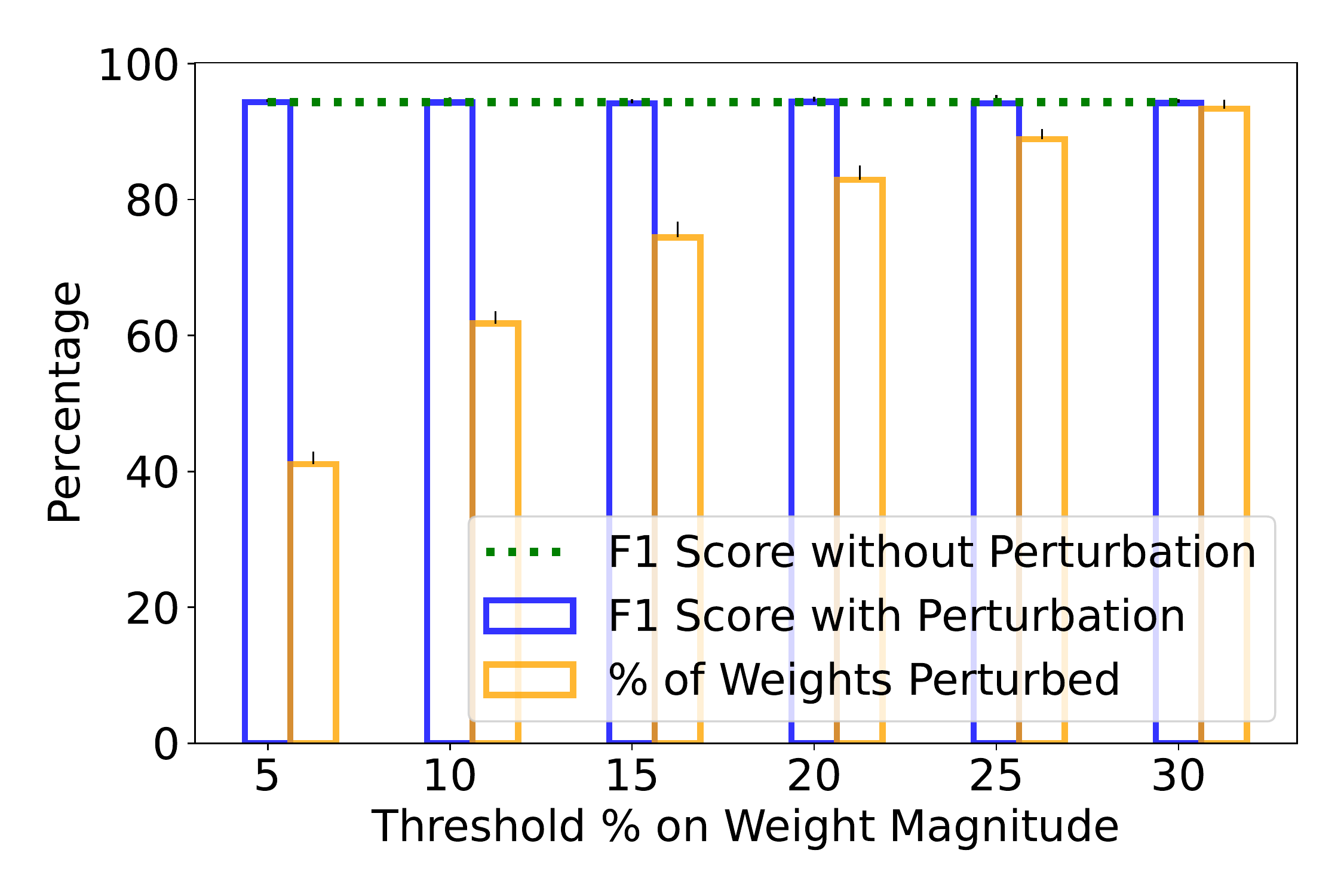}}
\subfigure[Parkinson's]{\includegraphics [width=0.32\textwidth]{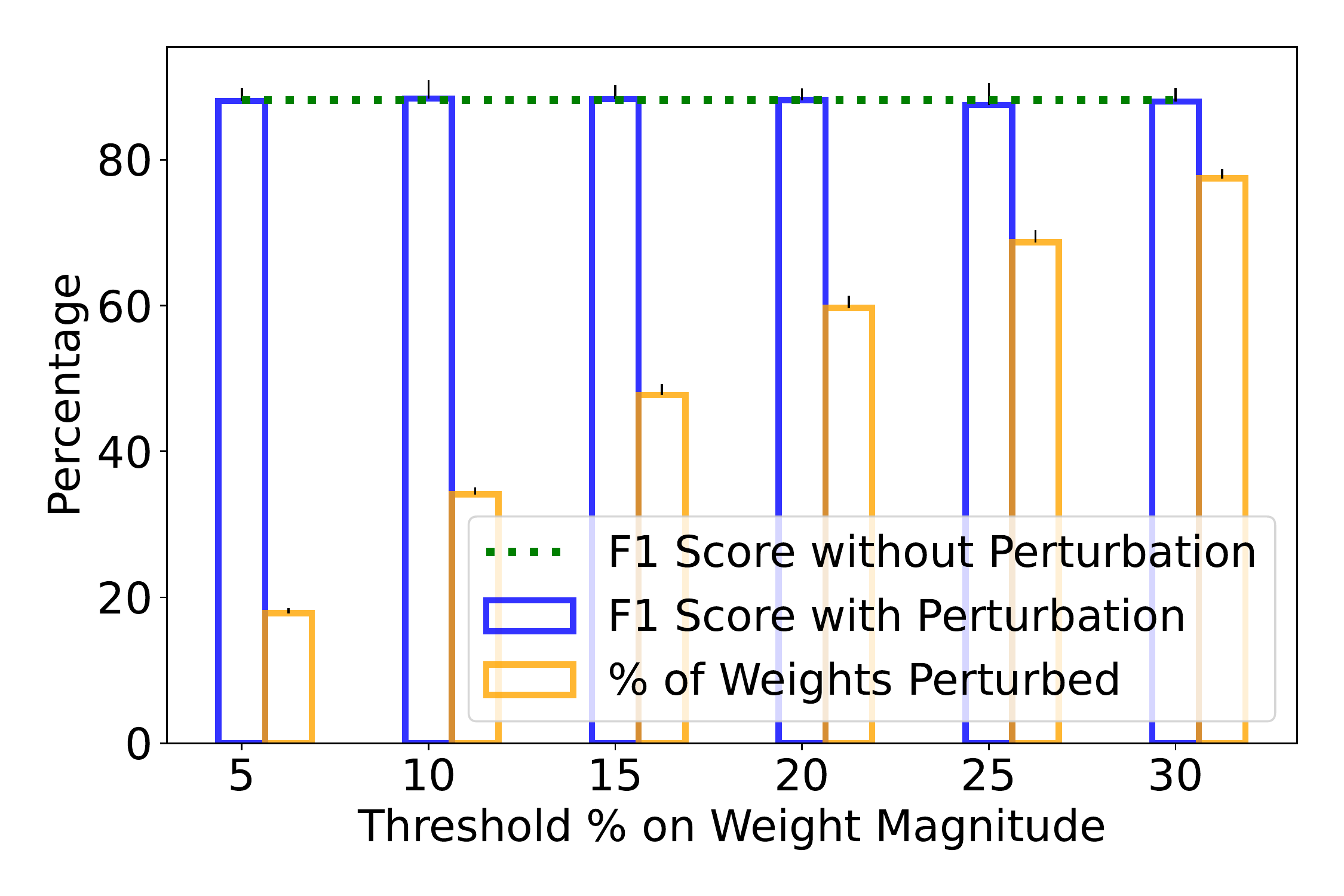}}
 	\vspace{-5pt}
\caption{The effect of perturbation parameter ($\tau$) on perturbing weights (\% of weights perturbed) in deep model pretraining and on downstream classification performance. Examples are shown for all six tabular data sets used in this study. }
\label{fig:ablation}
\end{figure*}

\subsection {Deep learning performance depends on data}

We explain the performance of deep models and traditional machine learning. One striking difference appears in the classification of the Madelon data set. Here, traditional machine learning (F1-score 81.6 (1.9)) substantially outperforms the best deep learning setup (F1-score 64.97 (0.8)). The correlation matrix of the Madelon data set suggests that the variables are highly uncorrelated with zero non-diagonal entries. This unusual result is because the Madelon data set is synthetically generated, unlike real-world datasets that inevitably contain correlated variables. Therefore, datasets with uncorrelated variables are poor candidates for deep learning. The high performance of deep models with gene sequence data (F1-score 99.75 (0.3)) explains why these models are so successful with sequential and correlated data (images, text).

\subsection{Ablation study} This study uses a default value of 5\% for the $\tau$ parameter to limit the model compression. A higher value of $\tau$ prunes a larger number of weights, which may hurt the downstream classification accuracy. This section investigates the effect of varying $\tau$ on model compression and downstream classification performance. Figure~\ref{fig:ablation} reveals the percentage of weights perturbed and F1-scores at downstream classification due to varying the perturbation parameter ($\tau$). As expected, increasing the $\tau$ value almost linearly increases the percentage of weights perturbed to introduce sparsity in the pretrained model. In the case of acrene and madelon data sets, a perturbation with $\tau$=25\% substantially improves the downstream classification F1-scores compared to the baseline model without such perturbation. In all other data sets, the downstream classification F1-scores remain stable, even after increasing the percentage of weight perturbation by increasing the value of $\tau$ during the pretraining. This observation can be explained by the notion that the weight pruning step is followed by the weight regrowing step to recover some of the lost weights by estimating the weight gradient on the pruned weights.

\subsection {Limitations}
Similar to any studies, this work has several limitations. The proposed weight perturbation does not have any positive effects on improving the performance of basic DNN models. Dropout still appears to be a powerful approach with basic DNN and datasets with uncorrelated variables. The improvement in F1-scores due to perturbation is between 0.2\% to 3\% across different data sets. This improvement may increase depending on the deep model, data set, and more importantly, the perturbation parameter $\tau$ as seen in the ablation study. Despite this limitation, weight perturbation additionally achieves 15\% to 40\% compression due to sparsity in the pretrained model. The proposed model compression is defined for the pretrained model, not at the finetuning or supervised classification stage. The general notion is that a generic pretrained model is stored and then finetuned to be repurposed for a target application. Therefore, sparsity in pretrained models enables compression for model storage and sharing.

\section {Conclusions}

This paper proposes a periodic weight perturbation routine while self-supervised pretraining of a deep neural network model. The proposed perturbation method can achieve both pretrained model compression and improved downstream classification accuracy across a variety of tabular data sets. This improvement often outperforms dropout learning and weight regularization when applied at the pretraining stage. A growing step followed by the pruning step can help correct some weights and learning trajectory, improving the downstream classification performance. Therefore, pruning or regularizing deep models at the self-supervised pretraining stage may be more effective than performing similar operations at the supervised training or finetuning step. Additionally, deep learning can perform very poorly in classifying tabular data sets with uncorrelated variables, whereas traditional machine learning achieves substantially high accuracy.

\section*{Acknowledgements}

Research reported in this publication was supported by the National Library Of Medicine of the National Institutes of Health under Award Number R15LM013569. The content is solely the responsibility of the authors and does not necessarily represent the official views of the National Institutes of Health.

\bibliographystyle{unsrt}
\bibliography{pruning}

\end{document}